%% file: Manuscript.tex
\documentclass[sn-chicago]{sn-jnl}

\jyear{2021}%

\theoremstyle{thmstyleone}%
\theoremstyle{thmstyletwo}%

\theoremstyle{thmstylethree}%

\usepackage{natbib}
\usepackage[section]{placeins}
\usepackage{comment}
\usepackage[title]{appendix}

\newcommand{\simpleheading}[1]{{\bf [#1]}}
\newcommand{\qstring}[1]{``{\it #1}''}
\newcommand{\pidlist}{\mathit{pid\textnormal{-}list}}
\newcommand{\sbp}{section-based partitioning}

\newcommand{\pbp}{paragraph-based partitioning}

\definecolor{applegreen}{rgb}{0.55, 0.71, 0.0}
\newcommand{\ming}[1]{{\color{applegreen} (Ming: #1)}}

\begin{document}
\title[Article Title]{Bringing Legal Knowledge to the Public by Constructing a Legal Question Bank Using Large-scale Pre-trained Language Model}









\author[1]{\fnm{Mingruo} \sur{Yuan}}\email{mryuan@cs.hku.hk}

\author[1]{\fnm{Ben} \sur{Kao}}\email{kao@cs.hku.hk}

\author[1]{\fnm{Tien-Hsuan} \sur{Wu}}\email{thwu@cs.hku.hk}

\author[2]{\fnm{Michael M.K.} \sur{Cheung}}\email{michaelmkcheung@hku.hk}

\author[2]{\fnm{Henry W.H.} \sur{Chan}}\email{hwhchan@hku.hk}
\author[2]{\fnm{Anne S.Y.} \sur{Cheung}}\email{anne.cheung@hku.hk}
\author[2]{\fnm{Felix W.H.} \sur{Chan}}\email{fwhchan@hku.hk}
\author[3]{\fnm{Yongxi} \sur{Chen}}\email{yongxi.chen@anu.edu.au}

\affil[1]{\orgdiv{Department of Computer Science}, \orgname{The University of Hong Kong}, \orgaddress{\street{Pokfulam}, \city{Hong Kong}, \country{China}}}

\affil[2]{\orgdiv{Faculty of Law}, \orgname{The University of Hong Kong}, \orgaddress{\street{Pokfulam}, \city{Hong Kong}, \country{China}}}

\affil[3]{\orgdiv{College of Law}, \orgname{The Australian National University}, \orgaddress{\street{ACT 2601}, \city{Canberra}, \country{Australia}}}
\abstract
Access to legal information is fundamental to access to justice. Yet accessibility refers not only to making legal documents available to the public, but also rendering legal information comprehensible to them. A vexing problem in bringing legal information to the public is how to turn formal legal documents such as legislation and judgments, which are often highly technical, to easily navigable and comprehensible knowledge to those without legal education. In this study, we formulate a three-step approach for bringing legal knowledge to laypersons, tackling the issues of navigability and comprehensibility.
First, we translate selected sections of the law into snippets (called CLIC-pages), each being a small piece of article that focuses on explaining certain technical legal concept in layperson's terms. Second, we construct a \emph{Legal Question Bank} (LQB), which is a collection of legal questions whose answers can be found in the CLIC-pages. 
Third, we design an interactive \emph{CLIC Recommender} (CRec).
Given a user's verbal description of a legal situation that requires a legal solution, CRec interprets the user's input and shortlists questions from the question bank that are most likely relevant to the given legal situation and recommends their corresponding CLIC pages where relevant legal knowledge can be found.
In this paper we focus on the technical aspects of creating an LQB. 
We show how large-scale pre-trained language models, such as GPT-3, can be used to
generate legal questions. We compare machine-generated questions (MGQs) against human-composed questions (HCQs) and find that MGQs are more scalable, cost-effective, and more diversified, while HCQs are more precise.
We also show a prototype of CRec and illustrate through an example how our 3-step approach effectively brings relevant legal knowledge to the public.





\maketitle

\input{1-intro.tex}

\input{2-relatedwork.tex}

\input{4-method.tex}

\input{5-evaluation.tex}

\input{6-application.tex}

\input{7-conclusion.tex}



\newpage
\bibliography{reference}

\begin{appendices}
\newpage
\input{appendix.tex}
\end{appendices}

\end{document}

%% file: 1-intro.tex
\section{Introduction}
\label{sec:intro}
With advances in information technology, legislation and judgments (i.e., the ``primary legal sources'' used by legal professionals), are available online for public accesses. Considerable attention has been given to online dissemination of legal information. Namely,
the {\it free access to law movement} (FALM)\footnote{\url{http://www.fatlm.org/}} is an international movement that created Legal Information Institutes (LIIs) around the world, each providing online databases of primary legal sources to the general public. 
The public availability of legal information in these formal sources, however, does not translate directly and easily
to legal knowledge to the public.
Although legal information can be freely retrieved, those without a legal education may still find the law to be 
``unreadable,'' difficult to understand, and thus not accessible, whether in common law countries~\citep{curtotti2015} or civil law jurisdictions~\citep[p. 53]{legal-nonlegal}.
There is generally what we coin as a {\it legal knowledge gap} between formal legal sources and the general public.
Identified by the New Zealand Law Commission and the New Zealand Parliamentary Counsel's Office~\citep{newzealand}, this knowledge gap is mostly attributable to three challenges:

\begin{description}
\item[{\bf Availability.}] The formal sources that express the law (including both legislation and case law) may not be easily {\it available} to the public in hard copy or electronic form. They may be difficult to locate or find.

\item[{\bf Navigability.}] The law expressed by the legal sources is hardly {\it navigable}, in the sense that the public may not have the ability to know of and reach the relevant legal rule or principle applicable to the situation they face.

\item[{\bf Comprehensibility.}] Even if the legal rule or principle is found, it is barely understandable, readable or {\it comprehensible} to the public. 
\end{description}

%

\begin{figure}
    \centering
    \includegraphics[width=0.9\textwidth]{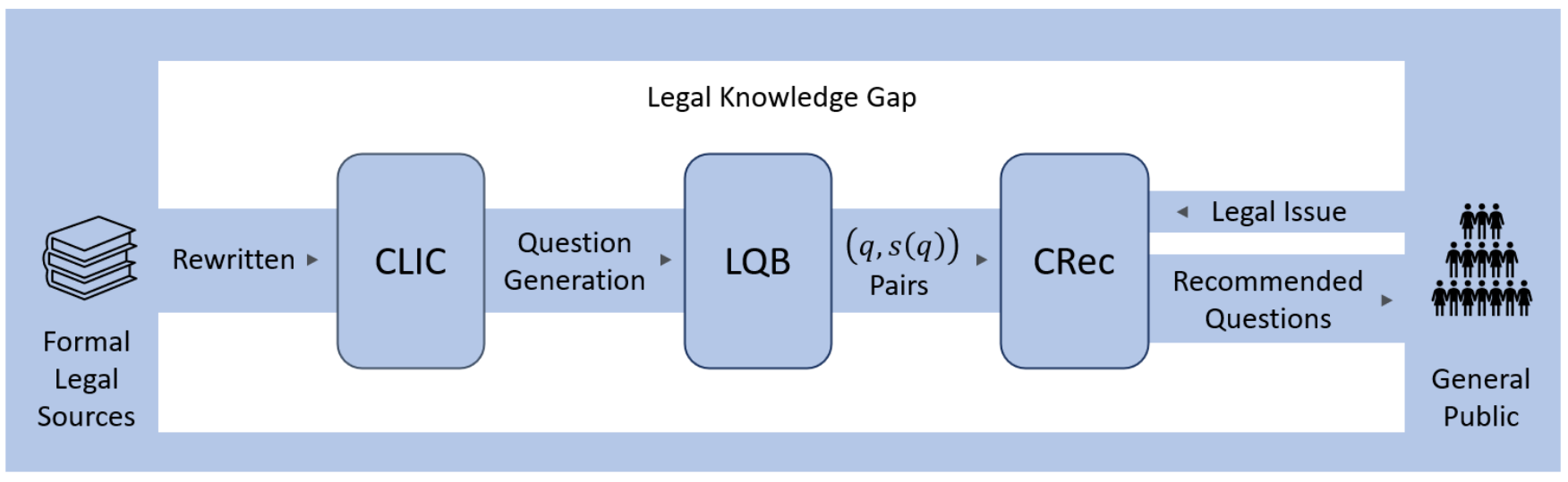}
    \caption{The legal knowledge gap and the three-step approach of bridging the gap.}
    \label{fig:overview}
\end{figure}

In recent years, great efforts have been made towards improving legal knowledge {\it availability}, particularly through online accesses of primary legal sources.
However, for people 
without formal legal training, it is often difficult for them to navigate the large volume of legal information, identify the correct legal issue, and find the relevant legal rules that they need. It is also challenging to understand the legal rules and their underlying concepts which determine the legal solution being sought. 
Hence, the objective of our study is to bridge the legal knowledge gap by addressing the problems of navigability and comprehensibility. 
Specifically, we investigate a three-step approach to bridging the gap. Figure~\ref{fig:overview} illustrates the legal knowledge gap and the three components of our approach. The components are further elaborated below.

\begin{figure}
	\centering
	\includegraphics[width=0.9\textwidth]{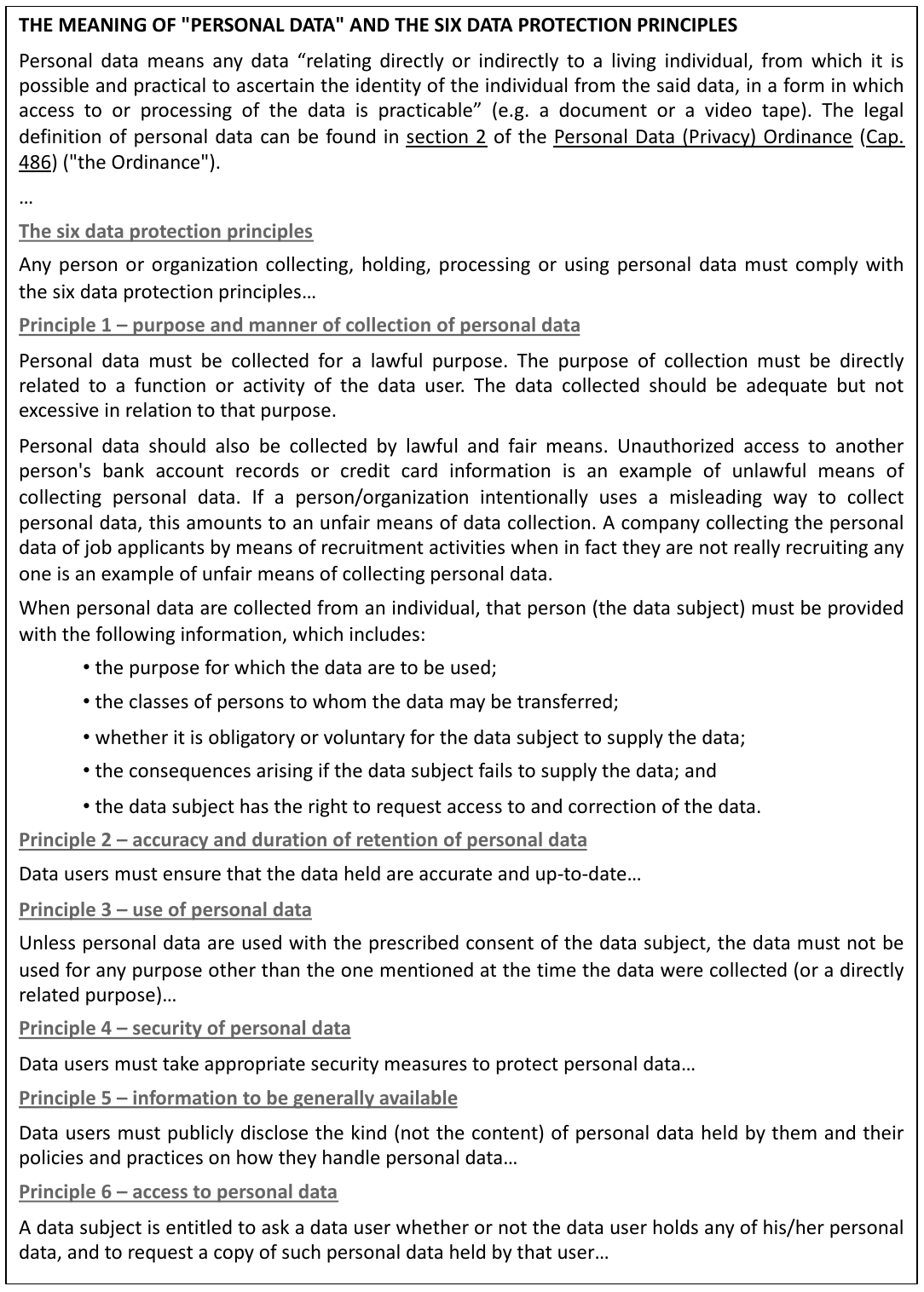}
	\caption{A CLIC-page excerpt (page id: 69).}
	\label{fig:page_example}
\end{figure}

\simpleheading{CLIC}
To tackle the issue of comprehensibility, the Law and Technology Centre of the University of Hong Kong has been running the Community Legal Information Centre (CLIC) since 2005\footnote{\url{https://www.clic.org.hk/}}.
It is an online platform that
provides a quick internet guide for the general public to find relevant legal information in Hong Kong. 
32 topics of the law that have the most direct bearing to people's daily lives are being covered. 
Some example topics include ``{\it Landlord and Tenant}'', ``{\it Traffic Offenses}'', ``{\it Medical Negligence}'', and ``{\it Sexual Offenses}''.
Technical legal concepts that are relevant to the selected topics are explained in layperson's terms and are presented in 
web pages on the CLIC platform. We call each such web page a {\it CLIC-page}. 
Essentially, each CLIC-page is a small article that focuses on explaining certain legal concept.
CLIC-pages are presented in different formats, such as illustrative cases and FAQs.
The objective of CLIC is to present the law (more precisely, the legal rules concerning given topics)
in less formal and less technical languages so that the law is more comprehensible to 
the general public. 
CLIC currently contains more than 2,000 CLIC-pages, which is to be expanded to 2,500--3,000 pages by the year 2024.
As an example,
Figure~\ref{fig:page_example} shows an excerpt of a CLIC-page, which explains six data protection principles.
Through this approach, the CLIC has directly addressed the issue of comprehensibility. However, the site is heavily dependent on a reader's skills in identifying the correct legal issue and finding the answer from the relevant web page. In other words, the ``navigability'' issue largely remains. 

\simpleheading{Legal Question Bank (LQB)}
To address the issue of navigability, and to improve comprehensibility of the law, 
the second component of our three-step approach is to construct a legal question bank (LQB) to enhance the human-legal interaction of the CLIC platform. 
The idea is to create a large collection of model questions whose answers could be found in some relevant CLIC-pages. 
For example, a model question for the CLIC-page shown in Figure~\ref{fig:page_example} is, 
\qstring{What are the points to pay attention to in collecting personal data?} 
This question can be answered by the paragraphs under ``Principle 1''  in the CLIC-page.
For each question $q$ collected in the question bank, we identify an answer scope $s(q)$, which indicates the CLIC-page and the paragraphs within which the answer to question $q$ can be found. ($q$, $s(q)$) thus forms a {\it question-answer} pair.
With the question bank, a user needs not express his/her legal question directly. Instead, the user can pick model questions from the question bank to phrase his/her legal concerns and retrieve their answers through the questions'  answer scopes.

\simpleheading{CLIC Recommender (CRec)}
As we will show later, dozens of questions can be created for each CLIC-page. 
With a target of collecting 100,000 questions into our LQB on CLIC, it is impractical for a user to identify relevant
model questions from the big LQB. 
Our third component is an AI assistant called the CLIC Recommender (or CRec). CRec is equipped with natural
language processing (NLP) capability such that it can converse with users and understand their legal issues.
The objective of CRec is to comprehend a user's verbal description of a legal scenario and shortlist a few model questions 
from the question bank that can most likely express the user's legal issues. 
Instead of sifting through the large question bank, the user only needs to go through a short list (say, 10) of questions
recommended by CRec. 

In this paper we focus on the technical aspects of creating the legal question bank. 
In particular, we investigate how questions and their answer scopes, that is, the ($q$, $s(q)$) pairs, can be effectively and 
efficiently created. We propose to apply a large-scale pre-trained language model, such as GPT-3, to create
{\it machine-generated questions} (MGQs). We also employ human workers to create
{\it human-composed questions} (HCQs). We compare and contrast the two methods of question creation. 
As we will see later in our evaluation results, MGQs are more scalable and cost-effective, and they are often more
diversified compared with HCQs. On the other hand, HCQs are generally more precise. 
Overall, the two question-creation methods complement each other well. 

The rest of the paper is organized as follows. 
Section~\ref{sec:relatedwork} discusses related works.
Section~\ref{sec:method} discusses the creation of the LQB. In particular, we give details on how to use
GPT-3 to generate MGQs. 
Section~\ref{sec:evaluation} gives a comprehensive evaluation of our question generation techniques. 
Section~\ref{sec:crec} discusses the design of CRec and illustrates it with an example. 
Finally, Section~\ref{sec:conclusion} concludes the paper.

%% file: 2-relatedwork.tex
\section{Context of the Problem}
\label{sec:relatedwork}
In this section we briefly explain the challenges to accessibility to law. Mentioned earlier, there are three major dimensions to the concept of accessibility. The first issue of availability is largely addressed by online dissemination of free legal information. While most literature has dealt with the issue of comprehensibility, the remaining issue of navigability has not been adequately studied and explored. 
We first present some existing works on the comprehensibility issue. 
Then, we introduce the pre-trained language model GPT-3 and other works on machine question generation,
which are designed to address the challenge of navigability and to enhance comprehensibility.

\subsection{Navigability and Comprehensibility of Legal Information}
Right of access to law is closely tied to the exercise of other fundamental rights such as  freedom of speech, access to justice and equality before law~\citep{Europeaccess}. Thus, attaining fair access to law is crucial to one's meaningful participation in a society. Similar to the opinion of the New Zealand Law Commission discussed earlier, \cite{Europeaccess} argues that the concept of access to legal information has embodied three different levels, namely, (1) making legal information available and searchable, (2) linking relevant legal information and documents, and (3) offering context dependent translation of legal knowledge to targeted audience. Based on this proposed standard, Mommers studies two legal database websites: EUR-Lex and www.officielebekendmakingen.nl, and finds that while the websites generally achieve the first two levels, the third one is largely ignored. It thus shows that ``comprehensibility'' of legal information to the general public is the most challenging issue.

In fact, comprehensibility of legal information is a perennial problem that scholars have been studying. 
\cite{readability2} conducts readability analysis of 407 text paragraphs in government legal services websites.
The analysis is done based on the readability standard established in the {\it The Plain Language Action and Information Network} (PLAIN)\footnote{https://www.plainlanguage.gov/}. 
The study shows that the readability of legal aid websites such as LawHelp.org and LawHelpInteractive.org is
generally beyond the comprehension of most Americans, especially those of vulnerable groups with lower literacy skills.
The study highlights the importance of appropriate readability standards of legal-aid information.
Also, 
\cite{readability} analyzes the readability of 201 legislations and related policy documents in the European Union (EU).
The analysis is done with five quantitative readability indices (Flesch-Kincaid, SMOG, ARI, Coleman-Liau, and Linsear Write).
It is found that  PhD-level education is required to comprehend certain laws and policy documents.
\cite{becher2021} uses the Flesch Reading Ease (FRE) test and the Flesch-Kincaid test to evaluate the readability of the privacy policies of 300 popular websites made to implement the EU General Data Protection Regulation (GDPR). Likewise, the study finds that most users found privacy policies largely unreadable.

This situation is regrettable since the ability to understand legal information is related directly to one's ability to ``navigate'' the legal matrix, i.e., the ability to identify the correct legal issue, to know and reach the relevant legal rule or principle, so as to solve one's legal problem. Many members of the public may not realize that their problems are of a legal nature and may not understand their legal rights concerning the problems. To empower the general public to access the law in its full dimension, we design an LQB, which helps the users to identify and capture the correct legal issue and relevant legal principle, to understand the law in plain legal terms, and to find the primary legal source.

\subsection{Question Generation and Pre-trained Language Models}
Here, we focus on the technical aspects of creating an LQB. In particular, we investigate how 
large-scale pre-trained language models (PLMs) such as GPT-3 are employed to create machine-generated questions (MGQs).
In this section, we give brief descriptions of works related to machine question generation and also those of PLMs.

There are a number of applications of machine question generation or recommendation. 
As an example, search engines can generate/recommend questions to users to help them better express their
search intents. To illustrate, if one googles \qstring{What is libel?}, the search engine will suggest 
a number of other (related) questions,
such as \qstring{What is the meaning of libel and slander?}. 
This helps users compose their queries with better choices of words and expressions, resulting in more effective search.
Machine question generation is also used in education, where students are asked (machine-generated) questions that
cover the contents of articles the students have read. This provides a low-cost approach to evaluate students' understanding of learning materials.
For example, \cite{case_study} presents a case study that shows how machine-generated short-answer questions can help
improve non-native English speakers' reading comprehension ability. 

Given a document $d$ and a span of text $a$ in $d$, the problem of answer-driven question generation is to generate questions $q$
whose answers are covered by (the intended answer) $a$. 
Example works on this problem include \cite{withans, withans2, withans3}. It is relatively straightforward to create a question $q$ if the answer $a$ is a simple factual statement by syntactic manipulation of the words in $a$. 
For example, from \qstring{John likes apples,} one can easily replace a noun by a question word to get \qstring{Who likes apples?}
The challenge lies in generating questions that go beyond simple single-sentence factoid questions to semantic-based 
questions that cover a larger span of text in the input. For example, given a document that lists the steps of litigation,
a good question would be \qstring{What are the phases of litigation?}
There are also works (e.g., \cite{multiagents}) that address the question-generation problem without assuming any intended answers (``$a$'')
are given. This is a more challenging problem because the machine is not guided by any given intended answers
and has to figure out the major focuses of the input text by itself. 
In our context, only CLIC-pages are given to the machine, and our machine question generation method has to 
create semantically-important questions that best summarize the information contained in the CLIC-pages. 

Early works on machine question generation employ rule-based methods to rewrite input sentences into questions through lexical and syntactic analysis~\citep{rule1,rule2,rule3}. However, these artificially created rules/templates are usually difficult to design and are not general enough to be applied across domains.
Recent works apply neural networks techniques to train question generation models. 
 \cite{du2017learning} designs the first fully data-driven approach considering question generation in a machine reading comprehension setting. \cite{multiagents} presents a multi-task learning framework to simultaneously identify question-worthy phrases in documents and construct questions accordingly. 
 A disadvantage of these neural-network-based approaches is that they require large sets of training data, which are 
very expensive to obtain. Instead, our approach is to use pre-trained models (PLMs) for unsupervised question generation.

Recent advances in PLMs see many interesting applications of unsupervised machine text generation. 
For example,
\cite{dino} proposes 
a method to obtain quality sentence embedding using PLM without data labeling, fine-tuning or modifying the pre-trained model. 
\cite{inpainting} proposes an interesting approach, called {\it dialog inpainting}, which converts a document (such as a Wikipedia page) into a synthetic two-person dialog. The idea is to treat sentences in a document as a writer's utterances and then
to predict what an imaginary reader would ask or say between the writer's utterances.
\cite{education_q} proposes a technique of generating questions using PLMs and how the technique is applied in education applications. 
The work studies prompting strategies used in executing the PLMs to generate questions given target answers.

GPT-3~\citep{gpt3} is a highly popular PLM that has an impressive text generation ability \citep{surveyforplm}.
GPT-3 is trained with a document collection that consists of around 500 billion words. The training data 
comes from multiple text sources, including Common Crawl \citep{common_crawl}, WebText2 \citep{web_text}, two book corpuses, and Wikipedia. The GPT-3 model has up to 175 billion parameters, which makes it 100 times larger than its
predecessor GPT-2 \citep{gpt2}. 
In this paper we will discuss how we leverage GPT-3's text generation ability to create MGQs.

%% file: 4-method.tex
\section{Method}
\label{sec:method}
In this section we describe our process of creating the legal question bank (LQB) from CLIC-pages. 
We first give a formal definition of the problem (Section~\ref{sec:definition}).  Then, we describe the steps we take 
to generate questions. These include (1) {\it Prompting} (Section~\ref{sec:prompt}), (2) {\it Partitioning} (Section~\ref{sec:partition}), and (3) {\it Deduplication} (Section~\ref{sec:deduplicate}).

\input{3-definition}

%

\subsection{GPT-3 Prompting}
\label{sec:prompt}
    GPT-3 \citep{gpt3} is a pre-trained language model with strong text generating capability. 
GPT-3 takes as input a {\it prompt} $\textbf{c} = [c_{0}, c_{1},... c_{M}]$, which is a sequence of word tokens
that serves as a ``context'' for text generation. 
Specifically, it can compute the probability of a given word appearing next in a textual sequence. 
Equation~\ref{equ:gpt} shows the process of auto-regressive generation, 
where $\theta$ represents GPT-3's model parameters, and 
$p_{\theta} (x_{t} \vert x_{0},...,x_{t-1},\textbf{c})$ gives the probability of generating 
$x_t$ as the $(t+1)$-st word given the prompt $\textbf{c}$ and the first $t$ words generated.
Given a prompt $\textbf{c}$, GPT-3 generates subsequent text probabilistically. 
    
    
    \begin{equation}
       p_{\theta}(\textbf{x} \vert \textbf{c}) = p_{\theta}(x_{0} \vert \textbf{c}) \prod_{t=1}^{T} p_{\theta} (x_{t} \vert x_{0},...,x_{t-1},\textbf{c}).
       \label{equ:gpt}
    \end{equation}
   
\begin{figure}
	\centering
	\includegraphics[width=0.9\textwidth]{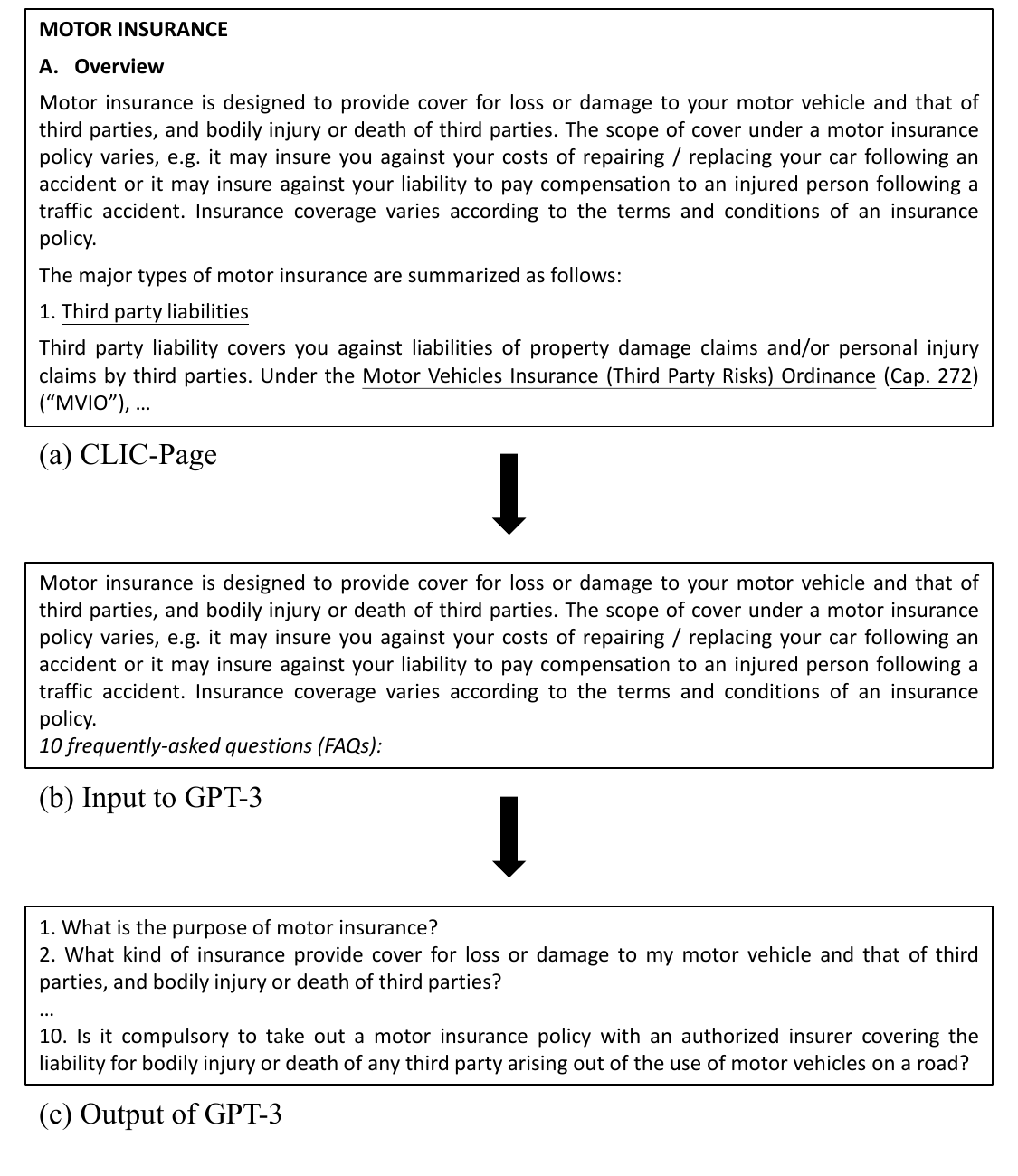}
	\caption{Prompting.}
	\label{fig:prompting}
\end{figure}

To induce GPT-3 to generate questions for a given CLIC-page $d$, we need to design a prompt $\mathbf{c}$ 
that provides the context of page $d$ and also instructs GPT-3 to output ``questions''. 
This is achieved in two steps.
First, we {\it partition} page $d$ into text segments. 
Second, for each text segment we append an FAQ sentence, literally, \qstring{10 frequently-asked questions (FAQs):} to the segment
to form a {\it prompt}. 
We then input each prompt (created for each segment of $d$) to GPT-3 to obtain questions.
All questions generated with respect to all the prompts are collected as questions of CLIC-page $d$.
Figure~\ref{fig:prompting} illustrates the process of constructing prompts from a CLIC-page.

In our study, we apply the following settings when we execute GPT-3:
$\mathit{temperature}$ = 1.0, $\mathit{top}\_p$ = 0.9, $\mathit{frequency}\_\mathit{penalty}$ = 0.3,
and $\mathit{presence}\_\mathit{penalty}$ = 0.1.

Next, we discuss different partitioning strategies, which result in different prompts created and thus different
question collections.

    
\subsection{CLIC-page Partitioning}
\label{sec:partition}
As we have mentioned, each CLIC-page could consist of multiple sections, each with multiple paragraphs.
Different sections and paragraphs focus on presenting different legal concepts, rules or principles, each considered a 
{\it legal knowledge element}. 
Ideally, the questions generated by GPT-3 align with the elements presented in a CLIC-page. 
Given a CLIC-page $d$, we consider the following partitioning strategies in prompt creation.

\simpleheading{Section-based partitioning}
Each section in $d$ forms one partition. We append the FAQ sentence, \qstring{10 frequently-asked questions (FAQs):} to each section to form one prompt. Each section of $d$ thus results in one collection of questions generated by
executing GPT-3 once. 

\simpleheading{Paragraph-based partitioning}
Each paragraph in $d$ forms one partition with the FAQ sentence appended to form one prompt. 
GPT-3 is executed once per paragraph to generate questions. 
Compared with section-based partitioning, paragraph-based partitioning creates more questions
(because GPT-3 is executed once per paragraph instead of once per section). 
However, the context given by each paragraph is narrower than that of each section. 
This may result in GPT-3 generating questions that are too specific (to each paragraph) instead of 
those that are more general and higher-level (that cover multiple paragraphs or a section).

\simpleheading{Hybrid partitioning}
The idea is to provide GPT-3 with section-level context but with paragraph-level attention. 
Specifically, we partition $d$ by sections. 
For each section, we prepend the paragraphs within it each with a paragraph label. 
A prompt is formed by appending the following sentence
\qstring{10 most frequently-asked questions for \texttt{[x]}}, where \textit{\texttt{[x]}} is a paragraph label.
We iterate through the paragraph labels of the section, and repeat the steps for each section in the CLIC-page $d$.
Under Hybrid partitioning, GPT-3 is executed once per paragraph in $d$ (as in paragraph-based partitioning)
and so the number of questions collected is comparable to that under paragraph-based partitioning.
Also, each prompt includes one section. This gives GPT-3 a broader context (as in section-based partitioning)
for more diversified question generation.

\begin{figure}
	\centering
	\includegraphics[width=0.9\textwidth]{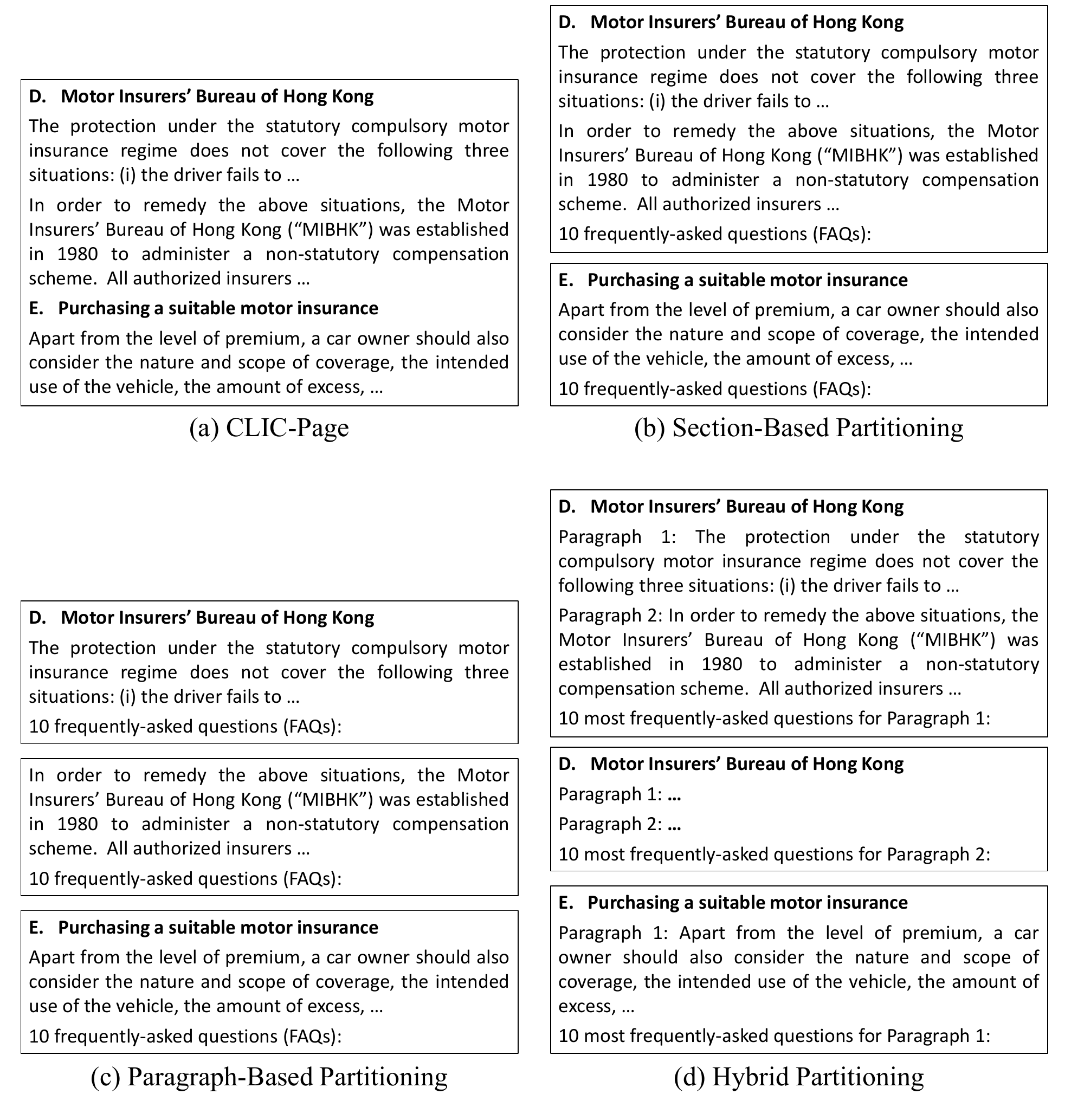}
	\caption{Partitioning Strategies.}
	\label{fig:partitioning}
\end{figure}

Figure~\ref{fig:partitioning} illustrates the three partitioning strategies in prompt creation.
We will compare the quality of the LQBs generated by the three strategies in Section~\ref{sec:evaluation}.

For each generated question $q$, we need to associate with it an answer scope $s(q) = [d: \pidlist]$ (see Section~\ref{sec:definition}).
We set $d$ to the CLIC-page from which $q$ is generated. Also, $\pidlist$ consists of all the paragraphs 
that are included in the prompt that generates $q$. 
The scope of a question can be further refined by manual verification.

\subsection{Deduplication}
\label{sec:deduplicate}
With paragraph-based partitioning and hybrid partitioning, GPT-3 is executed once to generate questions for each 
paragraph of a given CLIC-page. The process could generate tens of questions for a CLIC-page if the page is
long and content-rich. 
We observe that a small fraction of the questions generated by the model are redundant (i.e., they essentially ask the same 
legal questions with slightly different phrasing). We perform question deduplication to remove redundant questions in the LQB.

We first measure the similarity of questions in the LQB using sentence embedding. 
Specifically, we apply the pre-trained model DistilBERT\footnote{distilbert-base-nli-stsb-quora-ranking} to obtain an embedding vector
$v_q$ for each question $q \in$ LQB. 
For each pair of questions $q_1$ and $q_2$, we measure their similarity ($\mathit{sim}(q_1, q_2)$)
by the cosine similarity of their embedding vectors.
That is, 
\[ 
\mathit{sim}(q_1, q_2) = cosine(v_{q_1}, v_{q_2}) = (v_{q_1} \cdot v_{q_2}) / \lVert v_{q_1} \rVert \lVert v_{q_2} \rVert.
\]
We then apply single-link clustering to cluster the questions in the LQB with a similarity threshold of 0.95. 
For each cluster, we retain one question (chosen randomly) and discard the rest.

%% file: 3-definition.tex
\subsection{Problem Definition}
\label{sec:definition}
Let $\mathcal{D}$ be a collection of CLIC-pages. Our objective is to generate, for each document (CLIC-page) $d \in \mathcal{D}$, a set of questions $Q_d$. 
Specifically, each question $q \in Q_d$ is presented in a question-answer pair that takes the form
$(q, s(q))$, where $s(q)$ is called the {\it answer scope} of $q$. 
The scope $s(q)$ is represented by [$d$:~$\pidlist$], where $\pidlist$ is a list of paragraph id's in 
the CLIC-page $d$ where the answer of question $q$ can be found. 
Essentially, the answer scope $s(q)$ of question $q$ tells us which CLIC-page $d$ and which part of it 
(given by $\pidlist$) the question is answered. 
(We use $s(q).d$ and $s(q).\pidlist$ to represent the two pieces of information, respectively.)
In case a question $q$ generated is {\it irrelevant} to $d$ (i.e., $d$ does not answer $q$),
we set $s(q).\pidlist$ to NULL.
As an example, the question $q_1$: \qstring{What are the points to pay attention to in collecting personal data?} can be answered by paragraphs 3 to 5 of the CLIC-page shown in Figure~\ref{fig:page_example} (page id: 69). 
The answer scope of $q_1$ is therefore $s(q_1) = [69: \{3, 4,5\}]$.

We collect all the questions generated for all the CLIC-pages into an LQB. Here, we put forward
some desirable characteristics of a good-quality LQB.

\simpleheading{Quantity} 
The LQB serves as a collection of model questions for users to relate their legal situations. 
The LQB questions should therefore be of sufficient quantity so that the matching (between user's situation and 
model questions) is effective. 

\simpleheading{Precision}
The LQB should contain questions that are {\it relevant} to the CLIC-pages. That is, the questions in the LQB can be answered by the CLIC-pages.

\simpleheading{Coverage}
A typical CLIC-page consists of multiple paragraphs, each expressing a certain idea. 
Also, for more complex legal issues, their CLIC-pages are longer and may consist of multiple sections, each 
describing a certain aspect of a legal issue. 
A CLIC-page, therefore, often covers multiple facts or concepts. 
The questions generated for a CLIC-page should have good coverage of the page's contents.
That is, all of the paragraphs and sections should be included as part of the answer scopes of some questions generated.
In contrast, if the questions generated for a CLIC-page all cover, say, only the first paragraph of the page, then
we say that the questions have poor coverage of the page because the rest of the page is not addressed. 

\simpleheading{Diversity}
The questions generated for a page should have diverse scopes. 
That is, we want some questions that are {\it specific} that cover narrow scopes (e.g, over one paragraph of a page's content); we also want some questions that are of {\it higher conceptual level} that cover broader scopes (e.g., over a
section of a page or even the whole page). 

\simpleheading{Other Qualitative Measures}
In our study we found that some questions generated by GPT-3 for a CLIC-page are valuable and relevant
despite that the questions cannot be answered by the page's content. 
As an example, for the CLIC-page shown in Figure~\ref{fig:page_example} (on personal data),
GPT-3 generates the question 
\qstring{How can I make a complaint if I think my personal data has been mishandled?}
Note that the CLIC-page only explains what constitutes personal data mishandling; the page does not provide
information on the legal actions that a victim of such mishandling can take.
Although the question is not answered by the CLIC-page, it is nonetheless a very relevant question. 
Relevant but unanswered questions of this sort indeed help us identify areas of omissions, 
which give us hints on how certain CLIC-pages should
be revised and enriched. 
Moreover, the machine can occasionally generate multiple questions that essentially ask the same legal question 
but from different perspectives. 
For example, as we will illustrate later in Section~\ref{sec:results}, 
for the CLIC-page that addresses the legal issues related to ``Rates, Management Fees and other Charges'' under the landlord-and-tenant topic, 
the machine was able to 
generate different questions for the same details, some are from the perspective of tenants while others are from
the perspective of landlords. The different phrasing of the same question (from different perspectives) helps
to match different user's situations to model questions in the LQB more effectively, and provides different approaches to navigate the CLIC-page concerned.

In Section~\ref{sec:evaluation}, we will evaluate the machine-generated questions (MGQs) quantitatively with respect to the first four criteria. 
We will also give a qualitative study on the MGQs.

%% file: 5-evaluation.tex
\section{Evaluation}
\label{sec:evaluation}

\newcommand{\hquestion}[1]{\item[-] (manual) ``\textcolor{orange}{\it #1}''}
\newcommand{\mquestion}[1]{\item[-] (hybrid) ``\textcolor{purple}{\it #1}''}

\newcommand{\tabincell}[2]{\begin{tabular}{@{}#1@{}}#2\end{tabular}}

In this section we evaluate the performance of the three partitioning strategies and the quality of the LQBs they 
create with respect to the four quantitative measures, namely, {\it quantity}, {\it precision}, {\it coverage} and {\it diversity}
(see Section~\ref{sec:definition}).
We apply GPT-3 to generate questions for 1,557 CLIC-pages on the CLIC platform.
With the Hybrid strategy, we obtain 59,798 questions in the LQB. 

We manually evaluate machine-generated questions and compare them with human-composed questions. 
To do so,
we sampled 100 CLIC-pages under 5 topics on CLIC, namely, {\it landlord and tenant} (31), {\it defamation} (11), 
{\it insurance} (26), {\it personal data privacy} (9), and {\it intellectual property} (13).
(The numbers in parentheses give the number of CLIC-pages chosen for each topic.)
Specifically, we hire human workers, who have acquired formal legal training, to perform the following tasks:
\begin{enumerate}
\item Read the 100 CLIC-pages and compose questions whose answers can be found in those CLIC-pages. 
The workers are instructed to compose as many questions as they can. They are also required to identify
an answer scope for each question they create. Essentially, this process is to create an LQB manually for
a small set of 100 CLIC-pages. In total, the workers created 2,686 questions.
\item For each of the 100 CLIC-pages, read the machine-generated questions (MGQs) that are
generated by the three partitioning strategies. For each MGQ, 
determine if the question can be answered by the corresponding CLIC-page. If so, the question is marked 
``{\it correct}''. Also, the workers are asked to verify the answer scopes of the questions and modify them if necessary.
In total, the workers inspected 11,430 MGQs (that are generated by the three partitioning strategies).
\end{enumerate}

The above manual tasks took about 400 person-hours to complete. 
We remark that although the comparison study is performed over a small 100 CLIC-page sample, 
the study involves evaluating 11,430 MGQs and 2,686 human-composed questions (HCQs) for a total of about fourteen thousand questions.
The evaluation results we present in the rest of this section is therefore representative.

For illustration purpose, Appendix~\ref{sec:appendix} lists the questions generated by the various methods and by human workers for one of the CLIC-pages. 


\subsection{Results}
\label{sec:results}

{\bf Quantity.}
We start with comparing the number of questions generated by each strategy.
Recall that a larger LQB is more desirable because it provides more model questions to match against a user's legal
situation. This generally results in a higher success rate of helping a user to phrase his/her legal issue via the model questions. Moreover, a larger LQB is more likely to provide better coverage of the CLIC-pages' contents, 
and thus is more effective in capturing the legal knowledge presented in the CLIC-pages. 

Before we discuss the results, we would like to remark that while we instruct GPT-3 to generate 10 FAQs with each prompt, we do not set a hard quantity (10 questions per paragraph) for human workers. Instead, human workers are instructed to write ``as many questions as possible''. The reason is that we want to compare the two approaches (machine vs human) in a practical setting. Specifically, giving human workers a hard requirement of 10 questions per paragraph leads to several issues. First, human workers would interpret the instruction as to associate each question they compose to {\it one} specific paragraph. This would inadvertently cause workers to deliver {\it only} paragraph-level questions, i.e., those that are very specific to a paragraph, instead of more general questions that cover a wider scope (such as section-level questions). 
Secondly, the quantity of questions generated can vary based on the richness of information within a paragraph. Some paragraphs naturally yield a larger number of questions, while others do not. We opted to leave the decision of how many questions to generate per paragraph to the human workers to account for this variability.
Finally, it becomes progressively more difficult and time-consuming for human workers to write questions that are beyond what the workers naturally think are the most logical and interesting questions. Consequently, a hard 10-questions-per-paragraph instruction would significantly increase the time and cost of the manual question composition task.
For these reasons, we do not set a hard requirement (10 questions per paragraph) to human workers.

Table~\ref{tab:number} shows the number of questions generated by the three partitioning strategies for the sample of 100 CLIC-pages. 
Note that the questions given by each method have been deduplicated by our deduplication algorithm 
(see Section~\ref{sec:deduplicate}).
The number of human-composed questions (HCQs) is also shown (last column) as a reference. 
From the table, we see that \sbp\ gives  much fewer questions than 
\pbp\ or Hybrid. This is because the former executes GPT-3 once per CLIC-page section
to generate questions, while the latter two do that for each paragraph. GPT-3 is thus executed more times 
and generates more questions (almost four times as many) under \pbp\ and Hybrid compared
with \sbp. 
Moreover, \pbp\ and Hybrid generate similar number of questions. 
Comparing MGQs and HCQs, we see that machine is capable of creating more questions (about twice as many)
compared with human
workers, even though the human workers were instructed to make the best effort to compose as many questions as they 
could. 

\begin{table}[h]
\begin{center}
\caption{Number of questions created under different methods (after deduplication) (100 CLIC-page sample).}%
\begin{tabular}{|c|ccc||c|}
\hline
 & \multicolumn{3}{|c||}{MGQs} & HCQs \\
\hline
Method:  & Section-based  & Paragraph-based & Hybrid & Manual\\
\hline
Number of questions:   & 1,362   & 5,089  & 4,979 & 2,686 \\
\hline
\end{tabular}
\label{tab:number}
\end{center}
\end{table}

\begin{figure}
    \centering
    \includegraphics[width=0.9\textwidth]{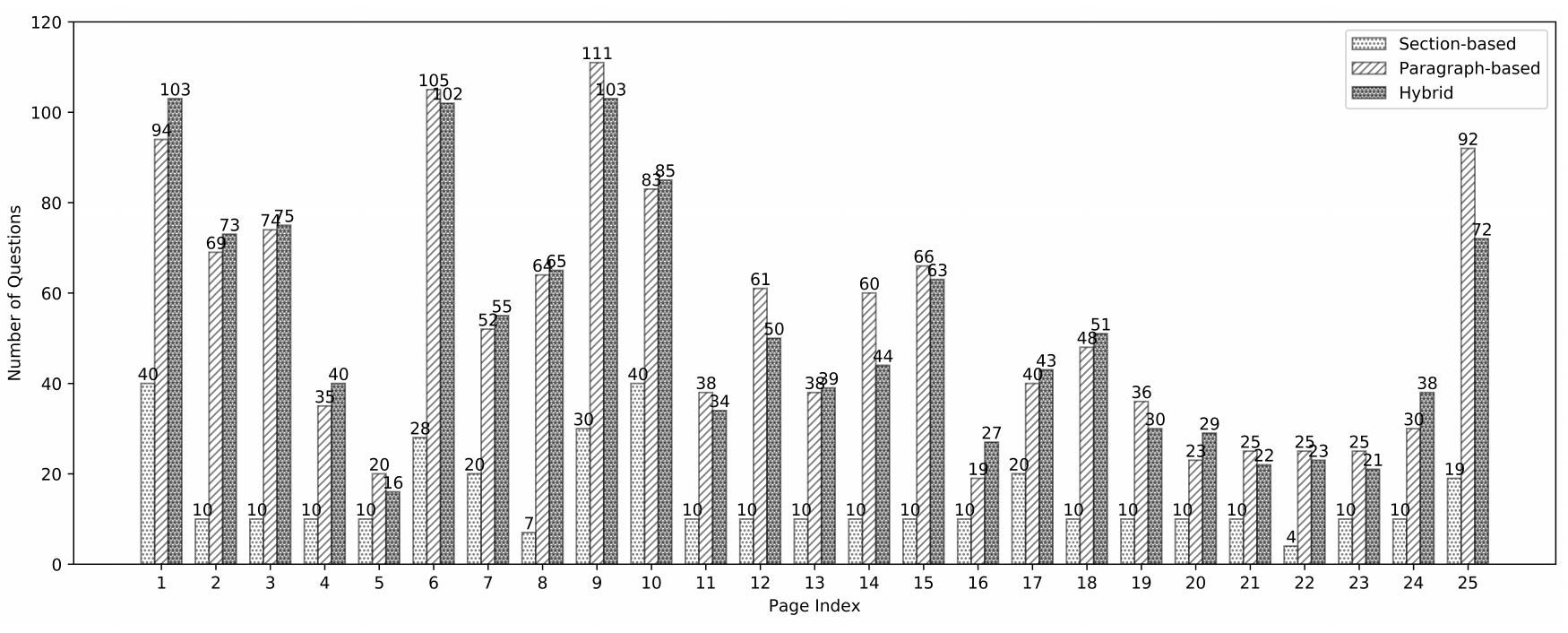}
    \caption{Number of  questions generated by each strategy (on 25 pages).}
    \label{fig:number}
\end{figure}

We further illustrate the differences among the partitioning strategies by showing the number of questions
generated by each of them for each of 25 example CLIC-pages that are considered in the evaluation. 
Figure~\ref{fig:number} shows the numbers. 
In the figure, the x-axis shows the CLIC-page id of the first 25 pages in our 100 CLIC-page sample. 
These 25 CLIC-pages come from two topics, namely, {\it landlord and tenant} (15 pages) and {\it defamation} (10 pages).
From the figure, we see that \pbp\ and Hybrid generally provide comparable numbers of questions, and they
consistently outperform \sbp\ by large margins. 
As a notable example, for CLIC-page-8, \pbp\ (64) and Hybrid (65) generate about 9 times as many questions as \sbp\ does (7). 
The reason is that CLIC-page-8 has a simple structure with one single section of 8 paragraphs. 
GPT-3 is therefore executed only once under \sbp, but many more times under \pbp\ and Hybrid.

\begin{figure}
	\centering
	\includegraphics[width=0.7\textwidth]{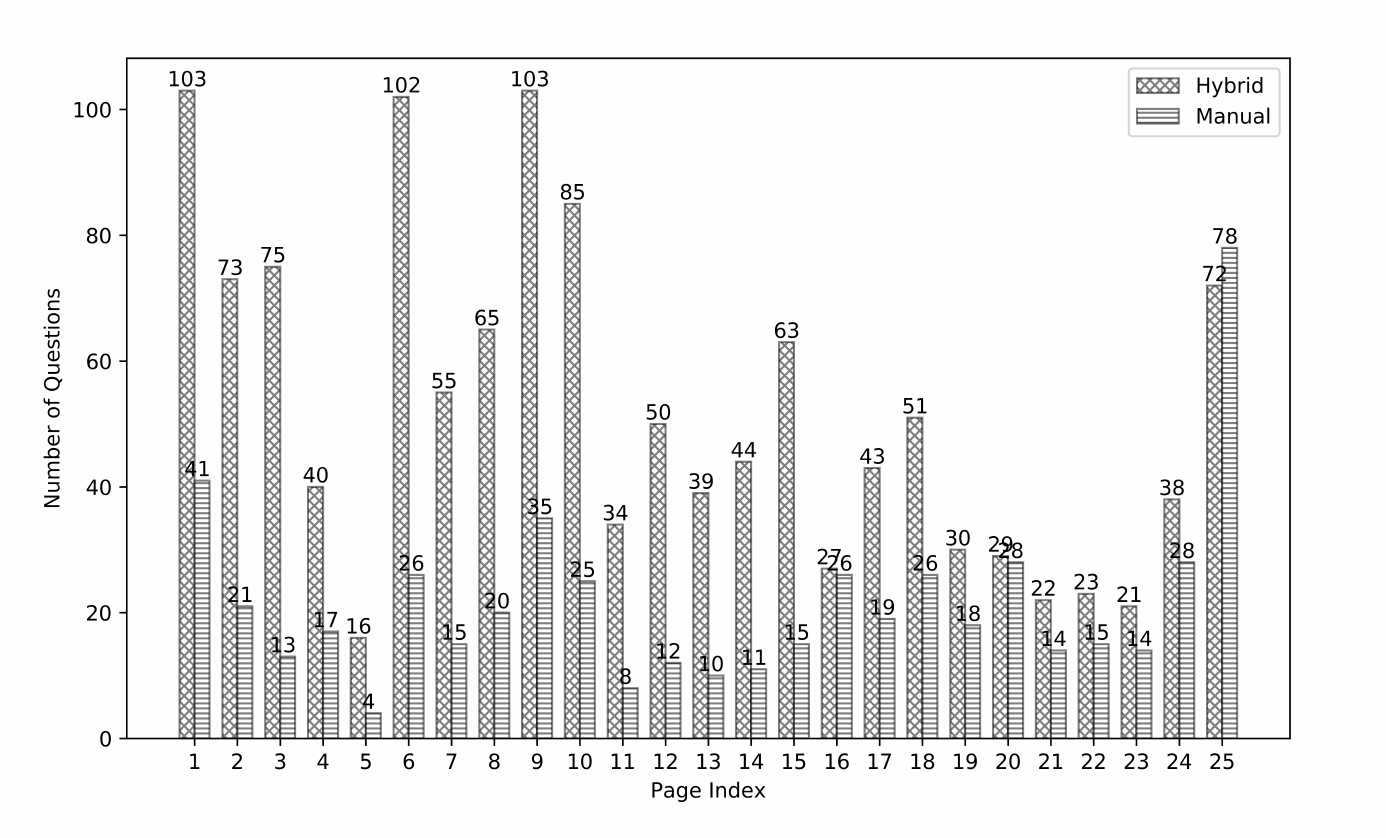}
	\caption{Number of questions generated; Hybrid vs human workers (on 25 pages).}
	\label{fig:number_human}
\end{figure}

Figure~\ref{fig:number_human} compares the number of questions generated by Hybrid and those by human workers
over the 25 CLIC-pages.
From the figure, we see that, in most cases, Hybrid generates much more questions than human workers. 
A representative example is CLIC-page-3, for which Hybrid gives 6 times as many questions as a human worker does.
This CLIC-page addresses the issue of property maintenance under the landlord and tenant topic.
We find that while human workers mostly ask questions from a tenant's perspective, 
the machine is able to ask questions from different perspectives, such as the perspective of a tenant as well as that of 
a landlord. Moreover, on the same piece of legal knowledge, the machine is able to ask questions on different details.
For example, part of CLIC-page-3 discusses a scenario in which a tenant is uncooperative when the landlord wants
to carry out maintenance work. 
On this legal issue, the machine asks multiple questions, such as
\qstring{Can the landlord apply for an injunction?}; \qstring{Can the landlord terminate the tenancy agreement?}
The machine generated questions are therefore richer with more variety. 

In conclusion, generally, we can obtain more MGQs than HCQs and with a much lower (human) cost. 
Among the three partitioning strategies, \pbp\ and Hybrid are much better in populating the LQB than \sbp.

{\bf Precision.}
Due to the probabilistic nature of GPT-3, questions generated by the machine for a CLIC-page are not guaranteed to be those 
that are answered by the CLIC-page's content. In our evaluation, every MGQ is verified by a human worker, who labels an MGQ $q$ as
``{\it correct}'' if $q$'s answer can be found in the CLIC-page $d$ from which $q$ is generated. 
We define the {\it precision} of a method as the fraction of questions the method generates that are labeled ``correct''.
Table~\ref{tab:accuracy} shows the precision of the three partitioning strategies when they are applied to the 100 CLIC-page sample.

\begin{table}[!tb]
\begin{center}
\caption{Question precision (over 100 CLIC-pages).}%
\begin{tabular}{|c|ccc|}
\hline
Method:  & Section-based  & Paragraph-based & Hybrid\\
\hline
Precision:   & 50\%   & 41\%  & 68\% \\
\hline
\end{tabular}
\label{tab:accuracy}
\end{center}
\end{table}

From Table~\ref{tab:accuracy}, we see that \pbp\ gives the lowest precision among the three strategies. 
The reason is that under \pbp, each prompt we input to GPT-3 (to generate questions) is a single paragraph of a CLIC-page. 
Since a single paragraph has limited content, the {\it context} (as given in the prompt) provided to GPT-3 is generally not rich enough
for it to create highly-precise questions. In contrast, \sbp\ improves precision over \pbp\ by providing richer contexts.
This is achieved by including a whole section of a CLIC-page in the GPT-3 prompt. 
Remarkably, Hybrid partitioning gives a much higher precision compared with the other two methods. 
The reason is that, like \sbp, Hybrid includes a whole section of a CLIC-page in prompting, which is context-rich. 
Also, Hybrid runs GPT-3 multiple times, each focusing on a specific paragraph (and thus a specific legal knowledge element) of a CLIC-page. 
This combination of {\it rich context} and {\it paragraph attention} is shown to allow GPT-3 to generate more precise questions. 

\begin{figure}
    \centering
    \includegraphics[width=0.9\textwidth]{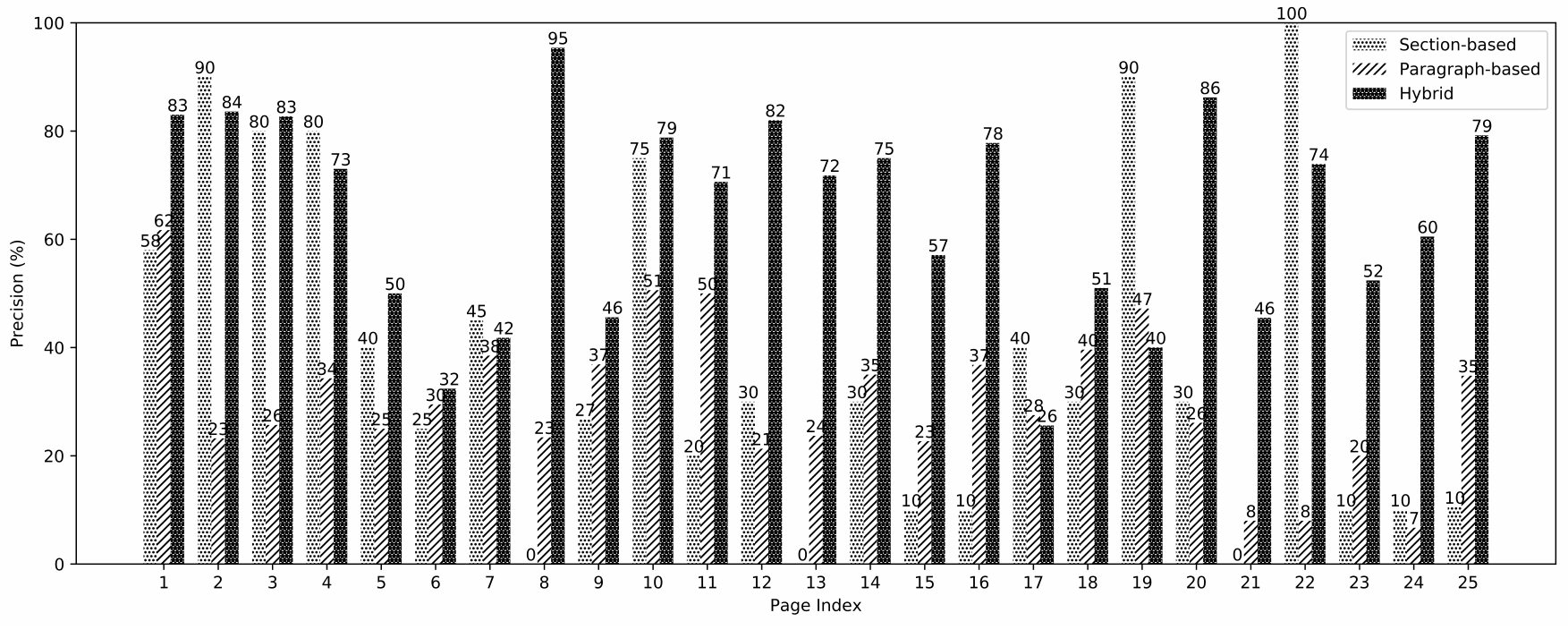}
    \caption{Methods' precision (on 25 pages).}
    \label{fig:accuracy}
\end{figure}

Figure~\ref{fig:accuracy} shows the precision of the methods when they are applied to each of the first 25 CLIC-pages of the 100 CLIC-page sample. 
From the figure, we see that the precision values (of each method) vary over the 25 CLIC-pages. 
This variation is due to the probabilistic nature of GPT-3. 
In general, we observe that Hybrid gives higher precision over the other methods.
In particular, Hybrid has the highest precision (compared with other methods) for 19 of the 25 CLIC-pages; 
Its precision exceeds 70\% for 14 CLIC-pages; Also, about 95\% of the questions Hybrid generates for CLIC-page-8 are correct.

In contrast, the performance of \sbp\ in terms of precision is less stable. From Figure~\ref{fig:accuracy}, we see that it gives
the best precision for 6 of the 25 CLIC-pages (CLIC-pages 2, 4, 7, 17, 19, and 22), but the worst precision for 13 other CLIC-pages.
In particular, all questions \sbp\ generates for CLIC-page-8 are incorrect (0\% precision), but the method generates only correct questions for CLIC-page-22 (100\% precision).

\begin{figure}
    \centering
    \includegraphics[width=0.9\textwidth]{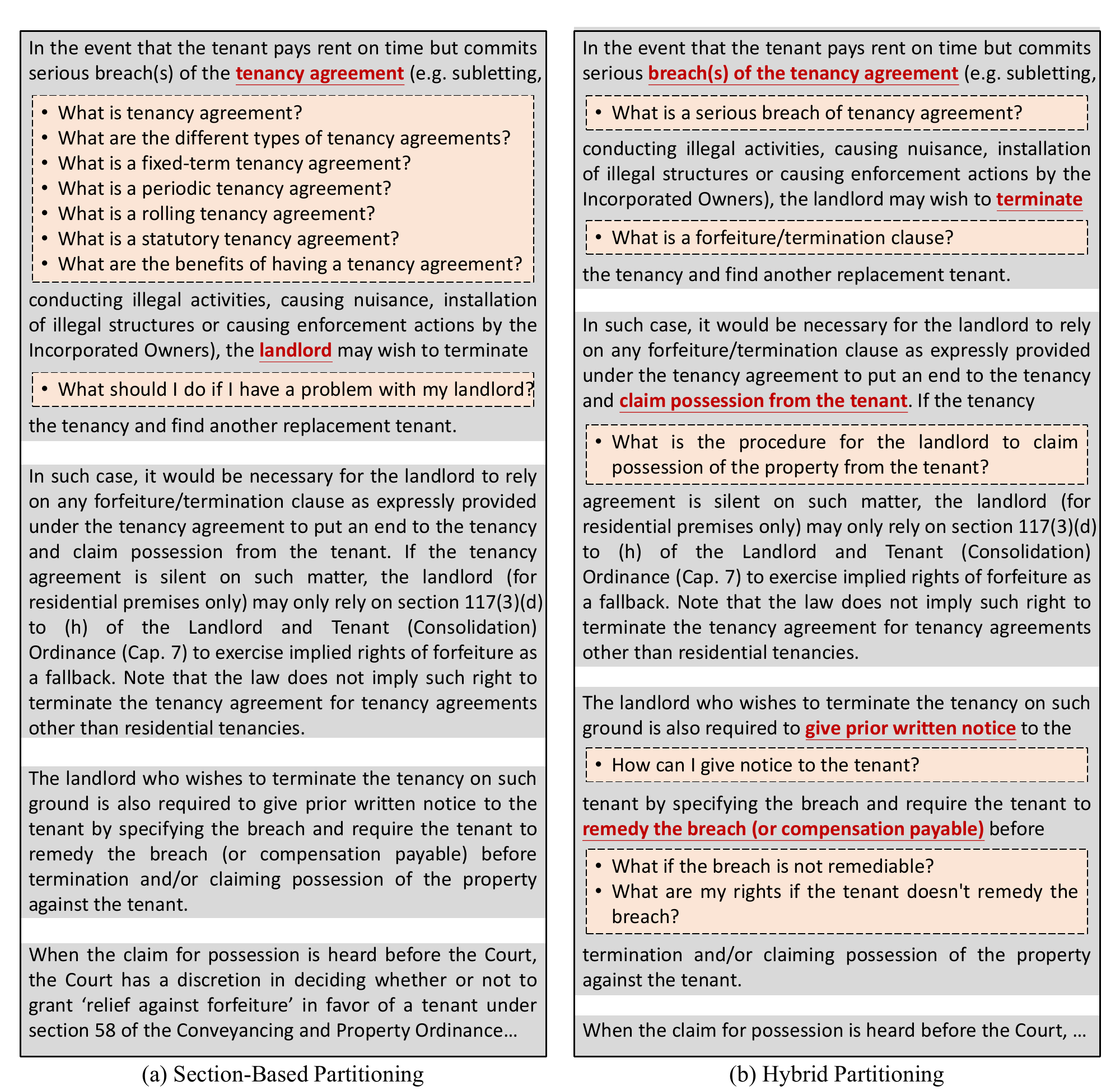}
    \caption{Illustration of focuses (underlined and highlighted in red) captured by GPT-3 and the corresponding questions generated (enclosed in dashed-boxes) with (a) section-based partitioning and (b) Hybrid.}
    \label{fig:focus}
\end{figure}

This variation in precision, especially for \sbp, is again due to the probabilistic nature of GPT-3. 
Given a context (prompt), GPT-3 will generate questions based on certain ``{\it focuses}'' it captures in the context. 
Figure~\ref{fig:focus} illustrates this observation. The left column of Figure~\ref{fig:focus} shows an excerpt of
a CLIC-page section, which consists of 4 paragraphs. 
The left column shows the result of applying \sbp, which submits the whole section to GPT-3 as the prompt (context).
In this example, GPT-3 captures two focuses  ``{\it tenancy agreement}'' and ``{\it landlord}''. 
Under each captured focus, we show the corresponding questions (enclosed in a dashed box) generated by GPT-3.
The right column of Figure~\ref{fig:focus} shows the same CLIC-page section but in this case GPT-3 is prompted 
by the Hybrid partitioning strategy. 
Similarly, in the right column, we show the focuses captured and the questions generated by GPT-3 
under Hybrid. 
Recall that Hybrid submits the whole section to GPT-3 as a prompt (like \sbp), but
repeats the process by iterating through the paragraphs taking one paragraph as the attention at a time
(see Figure~\ref{fig:partitioning}).  
From Figure~\ref{fig:focus}, it is evidenced that Hybrid is able to capture important focus in each paragraph. This results in 
more precise questions.

We remark that whether the questions generated by GPT-3 are correct or not depends on whether GPT-3 is
able to successfully capture meaningful focuses in the context.
For example, the CLIC-page section shown in Figure~\ref{fig:focus} is about the legal issues
concerning breaches of a tenancy agreement.
When GPT-3 is executed once on the section under \sbp\ (left column), the focuses GPT-3 captures, such as ``tenancy agreement'', are not very precise. The questions generated in this case are therefore not quite correct.
However, under Hybrid, GPT-3 is given, not only the whole section as a context, but also specific attention to the
paragraphs, one at a time. As is seen in Figure~\ref{fig:focus}'s right column, GPT-3 is able to capture focuses
more precisely, resulting in more precise questions. 
 

%

So far, we have seen two advantages of Hybrid compared with \sbp\ and \pbp: 
It gives large number of questions (Table~\ref{tab:number}) and more precise questions (Table~\ref{tab:accuracy}).
So, based on the measures of {\it quantity} and {\it precision} combined, Hybrid is a better strategy. 
An interesting question is how Hybrid fares against human workers in constructing an LQB. 
For the 100 CLIC-page sample, Hybrid generates 4,979 questions. 
At a precision of 68\%, we have 4,979 $\times$ 68\% $\approx$ 3,400 correct questions. 
This compares favorably against human-composed questions:
Not only does Hybrid give more (correct) questions (3,400 MGQs vs 2,686 HCQs),
it is also a more cost-effective approach. 
The latter is because it is much easier and faster to manually verify an MGQ than to manually create an HCQ. 

{\bf Coverage and Diversity.}
In Section~\ref{sec:definition}, we define the scope $s(q)$ of a question $q$ generated from a CLIC-page $d$ as
$[d: \pidlist]$, where $\pidlist$ is a list of paragraph id's from which the answer of question $q$ can be found.
We say that question $q$ {\it covers} the contents presented in the paragraphs listed in $\pidlist$.
For a good LQB, the question bank should contain sufficient questions so that all contents of the CLIC-pages are covered.
We measure the {\it coverage} of an LQB by the fraction of paragraphs in the CLIC-page collection that are covered by the questions
in the LQB. Formally,
\begin{equation*}
\mathit{coverage}(\mathit{LQB}) = \left\lvert \underset{q \in \mathit{LQB}}{\cup} s(q).\pidlist \right\rvert / N_p,
\end{equation*}
where $N_p$ is the total number of paragraphs in the CLIC-page collection.

As an illustration, there are in total $N_p = 599$ paragraphs in the 100 CLIC-page sample.
Table~\ref{tab:coverage} compares the coverages of the LQBs created by the three partitioning strategies
and also the one created by human workers.

\begin{table}[h]
\begin{center}
\caption{Coverages of LQBs created by different methods (100 CLIC-page collection).}%
\label{tab:coverage}
\resizebox{\linewidth}{!}{
\begin{tabular}{|c|ccc||c|}
\hline
 & \multicolumn{3}{|c||}{MGQs} & HCQs \\ \hline
Method:  & Section-based  & Paragraph-based & Hybrid & Manual\\\hline
Number of paragraphs covered: & 312  & 422 & 557 & 587  \\ \hline
Coverage:   & 52.1\%   & 70.5\%  & 93.0\% & 98.0\% \\ \hline
\end{tabular}
}
\end{center}
\end{table}

From Table~\ref{tab:coverage}, we see that \sbp\ has a relatively poor coverage (52.1\%). 
The LQB \sbp\ created covers only about half of the CLIC-pages' contents. 
In contrast, Hybrid partitioning gives a very high coverage of 93\%.
The reason why Hybrid gives a much higher coverage compared with \sbp\ is that
Hybrid uses the whole section as context in prompting (similar to \sbp) but it 
repeats executing GPT-3, each time with a paragraph attention. 
These two factors combined allows GPT-3 to capture important focus from each paragraph 
so that it can generate precise questions for each of them. 
Figure~\ref{fig:focus} clearly illustrates this observation. 
The left column in the figure shows that \sbp\  generates questions only for the first paragraph
of the CLIC-page (and the focuses it captures are not precise, leading to incorrect questions).
Hybrid (right column), in contrast, is able to lead GPT-3 to capture meaningful focuses (and hence
correct questions) for each paragraph. 
This contributes to Hybrid's high-coverage performance. 
The better coverage by Hybrid over paragraph-based partitioning is further illustrated by the example
shown in Appendix~\ref{sec:appendix}. From the answer scopes of the questions listed in the appendix, we see that
Hybrid gives questions that cover all eight paragraphs of CLIC-page $p_3$, while paragraph-based partitioning
fails to cover three of the eight paragraphs. 

From Table~\ref{tab:coverage}, we also see that Hybrid outperforms \pbp\ (whose coverage is 70.5\%). 
The reason is that although \pbp\ executes GPT-3 for each paragraph, each prompt \pbp\ uses
consists of just one paragraph. It gives insufficient context to GPT-3 to generate correct questions, leading
to \pbp's poor precision (see Table~\ref{tab:accuracy}). 
As a result, even \pbp\ generates questions for each paragraph, many of those questions are not correct
and thus their scopes do not cover the paragraphs. 

Comparing Hybrid against human workers, we see that Hybrid's performance approaches that of HCQs in terms
of coverage. We remark that the human workers were instructed to compose questions that cover all the contents
of the CLIC-pages. It is highly impressive that the LQB generated by machine (Hybrid) can achieve human-like
performance.

Next, we compare Hybrid against human workers in terms of question diversity. 
We argue that a desirable characteristic of a good LQB is that it contains {\it diversified} questions. 
By diversity, we mean 
(1) there are questions that ask for the same piece of legal knowledge from different perspectives,
and 
(2) there are questions of different levels of specificity/generality. 
Let us illustrate question diversity by comparing the questions
generated by Hybrid and human workers, respectively, for CLIC-page-8.

\begin{figure}
    \centering
    \includegraphics[width=0.7\textwidth]{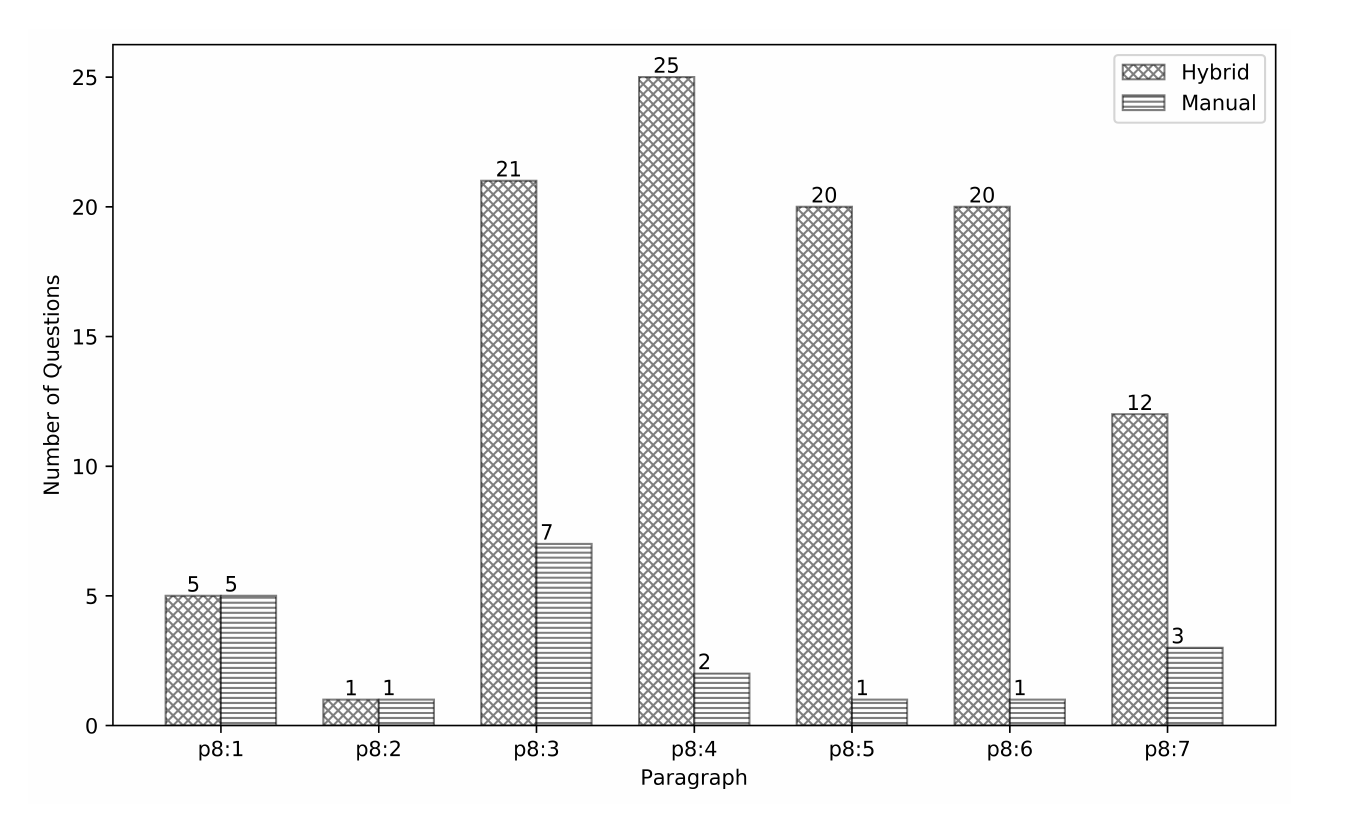}
    \caption{Number of questions created by Hybrid and human worker for each paragraph in CLIC-page-8.}
    \label{fig:coverage_finer}
\end{figure}

From Figure~\ref{fig:number_human}, we see that the numbers of questions generated for CLIC-page-8
by Hybrid and human workers are 65 and 20, respectively.
There are 7 paragraphs in CLIC-page-8. We use the notation $p_8$:$x$ to refer the $x$-th paragraph of page-8.
Figure~\ref{fig:coverage_finer} shows the number of questions that cover each paragraph
of CLIC-page-8 by the two methods.
For example, there are 21 questions given by Hybrid that cover paragraph 3 while there are 7 such questions given 
by human workers.
If one adds up the quantities of the bars (in Figure~\ref{fig:coverage_finer}) given by each of the methods,
we get 104 (Hybrid) and 20 (Manual).
Note that the tally 104 for Hybrid is much larger than the number of questions Hybrid generated (65).
This is because many questions given by Hybrid have scopes that cover multiple paragraphs. 
These {\it multi-paragraph questions} are those that are of higher-level of generality, which require multiple 
paragraphs to answer.
For example, question Q49 given by Hybrid (see Appendix) is a multi-paragraph one, which covers both paragraphs 5 and 7 of the CLIC-page.
In contrast, the bar-quantity tally for Manual (20) is exactly the same as the number of questions Manual 
generated. That means each question composed by the human worker covers only one single paragraph.
These questions are very specific to the legal knowledge carried by each paragraph. 
From this observation, we see that the machine (Hybrid) is able to create more diversified questions (from general to 
specific), while human workers tend to compose mostly specific questions.

We further illustrate question diversity by showing the questions created for Paragraph 6 (i.e., $p_8$:6).
Page-8 is about ``Rates, Management Fees and other Charges''. 
For paragraph 6, the human worker gave only one question while Hybrid generated multiple (see Figure~\ref{fig:coverage_finer}). In Figure~\ref{fig:example1}, we show the only question that was created manually
and some of the questions generated by Hybrid. 
We see that the questions given by Hybrid are richer and more diversified. 
In particular, the questions cover different user perspectives such as landlord's, tenant's, and even manager's
and co-owners'. 

\begin{figure}[h]
\resizebox{\linewidth}{!}{
	\scriptsize
\begin{tabular}{@{}l@{}}
\toprule
 \tabincell{c}{
 Page $p_{8}$: ``RATES, MANAGEMENT FEES AND OTHER CHARGES'' 
\\\url{https://clic.org.hk/en/topics/landlord_tenant/ratesManagementFees}}\\

\midrule

(manual) \textcolor{teal}{\it \tabincell{l}{Who is liable for the management fees or other forms of contributions \\(e.g. renovation costs and contribution to litigation funds) to \\be made pursuant to the Deed of Mutual Covenant (“DMC”)\\ or the Building Management Ordinance (Cap. 344) (the “BMO”)?}}\\

\midrule

(hybrid) \textcolor{purple}{\it \tabincell{l}{What happens if a tenant defaults on payment of management fees or \\other contributions to be made pursuant to the Deed of Mutual Covenant \\(“DMC”) or the Building Management Ordinance (Cap. 344) (the “BMO”)?}}\\
\midrule

(hybrid) \textcolor{purple}{\it \tabincell{l}{What is the landlord's liability for payment of management fees or\\ other forms of contributions pursuant to the Deed of Mutual Covenant \\(DMC) or the Building Management Ordinance (BMO)?}}\\
\midrule

(hybrid) \textcolor{purple}{\it \tabincell{l}{Is an obligation to pay money under the DMC/BMO directly enforceable \\by the manager/incorporated owners against a tenant?}}\\
\midrule

(hybrid) \textcolor{purple}{\it \tabincell{l}{Can the manager or other co-owners enforce the payment of management \\fees directly against a tenant?}}\\
\midrule

(hybrid) \textcolor{purple}{\it \tabincell{l}{Can a tenant still be held liable for any payment defaults even if the \\lease/tenancy agreement provides that the tenant shall make \\payment directly to the management office?}}\\

\botrule
\end{tabular}
}
\caption{Example questions created for $p_8$:6.}
\label{fig:example1}
\end{figure}

\subsection{Further Discussion}
In the previous section we compared the three partitioning strategies quantitatively based on the LQBs 
created by them in terms of {\it quantity}, {\it precision}, {\it coverage}, and {\it diversity} measures. 
We showed that Hybrid partitioning outperforms the other partitioning schemes, and that applying GPT-3 with
Hybrid gives machine-generated questions (MGQs) that compare favorably
against human-composed questions (HCQs).
In this section we further discuss some advantages of MGQs over HCQs and give illustrations to our findings.

{\bf Augmenting questions.} 
In Table~\ref{tab:accuracy}, we showed that Hybrid gives an LQB of 68\% precision over the 100 CLIC-page sample.
That means that given an MGQ $q$ generated for a CLIC-page $d$, 
there is a 32\% chance that the page $d$ does not directly answer question $q$ 
(and so $q$ is not labeled ``correct'' by our human verifier). 
When constructing an LQB, these  ``unanswered'' MGQs should be removed. 
We took a closer look at the unanswered MGQs and we found that many of these MGQs are indeed 
interesting questions that are relevant to the legal issues discussed in the CLIC-pages. 
They are not labeled ``correct'' simply because the CLIC-pages are not comprehensive enough 
to cover the legal issues raised by the questions. 
We call this kind of machine-generated questions, which are interesting and relevant to their CLIC-pages but
are not answered directly by the pages because of insufficient contents in the pages, {\bf augmenting questions}.
Figure~\ref{fig:example2} gives two example augmenting questions.
The first example concerns CLIC-page-5, which discusses legal issues when a landlord sells a property with an 
existing tenancy. 
An augmenting MGQ, $q_a$, reads \qstring{What if the tenant refuses to allow access for viewings?} 
Note that question $q_a$ addresses the legal rights of the landlord in case the tenant is uncooperative and
is obstructing the sale. CLIC-page-5, however, only states what the landlord needs to do to sell his property (such as
the parties the landlord needs to inform). Therefore, question $q_a$ is not labeled ``correct'' by our human verifier.
Question $q_a$, however, is very relevant to the legal topic of the CLIC-page. 
The question provides hints on how the content of the CLIC-page could be enhanced (e.g., by explaining the
legal rights of the landlord).
The second example in Figure~\ref{fig:example2} concerns CLIC-page-21, which is about defamatory articles. 
An augmenting MGQ, $q_b$, we obtained from Hybrid states a highly-related question on defamatory remarks
in private conversation. Although question $q_b$ is not answered directly by CLIC-page-21, the MGQ again
suggests how the content of the page can be enriched.

\begin{figure}[h]
\scalebox{0.75}{
\begin{tabular}{@{}l@{}}
\toprule
 \tabincell{l}{
 Page $p_{5}$: ``LANDLORD SELLS THE PROPERTY WITH EXISTING TENANCY'' 
\\\url{https://www.clic.org.hk/en/topics/landlord_tenant/landlordSellsThePropertyWithExistingTenancy}}\\
\midrule
{\it \tabincell{l}{(Augmenting) $q_a$: What if the tenant refuses to allow access for viewings?
}}\\
\botrule
\botrule

 \tabincell{l}{
 Page $p_{21}$: ``IF I DID NOT INTEND TO REFER TO THE PLAINTIFF IN MY ARTICLE, \\AND IT WAS A PURE COINCIDENCE THAT THE ARTICLE APPEARS TO REFER TO HIM, \\WILL I STILL BE LIABLE FOR DEFAMATION?'' 
\\\url{https://clic.org.hk/en/topics/defamation/identifying_the_person_defamed/q1}}\\
\midrule
{\it \tabincell{l}{(Augmenting) $q_b$: If I make a statement in private conversation, can I still be liable for defamation?}}\\
\botrule
\end{tabular}
}
\caption{Example augmenting questions generated by Hybrid.}
\label{fig:example2}
\end{figure}

We remark that augmenting questions, even though they are labeled ``incorrect'', should be retained in the LQB
and should be used to guide CLIC-page revision. 
We manually inspect all MGQs created by Hybrid on the 100-CLIC-page sample 
to identify all the augmenting questions it generates.
We find 69 augmenting questions covering 37 (of the 100) CLIC-pages. 
This is substantial because the discovery suggests how nearly 40\% of the CLIC-pages can be enriched. 
Human workers on the other hand create only ``correct'' questions; We do not observe any augmenting questions in the HCQs.

{\bf Question diversity and generality.}
In the previous section we showed that the MGQs obtained with Hybrid are more diversified than HCQs. 
In particular, some MGQs have answer scopes that span multiple paragraphs, while HCQs are usually focusing on
each individual paragraph. 
We argued that questions whose answer scopes include multiple paragraphs (or {\it ``multi-paragraph questions''} for short)
are higher-level ones and are more {\it general} in the sense that
they require more elaborate contents (more paragraphs) to answer. 
In fact, users without legal training tend to ask more general questions because they often do not know 
the relevant legal issues well enough to ask very specific (single-paragraph) questions.
As an illustrative example, CLIC-page-12\footnote{\url{https://clic.org.hk/en/topics/landlord_tenant/terminationOfTenanciesByNonpaymentOfRent}} addresses the issue of ``termination of tenancies by non-payment of rent''. We inspected the questions generated by Hybrid. Two example questions, one general and another specific, are:
\begin{itemize}
    \item (General question) $q_c$: ``{\it I am a landlord and I want to evict my tenant who has not been paying rent. Can I do so?}''; Answer scope: [$p_{12}$: \{1, 2, 3, 4\}].
    \item (Specific question) $q_d$: ``{\it Can a tenant be evicted if he has been granted relief against forfeiture?}'';
    Answer scope: [$p_{12}$: \{6\}].
\end{itemize}
Note that the general question $q_c$ has a scope that spans the first four paragraphs of CLIC-page-12, while the
specific question $q_d$ has its answer given in a single paragraph (paragraph 6 of the CLIC-page).

We inspected all the MGQs given by Hybrid and all the HCQs to label each question as 
{\it general} (whose answer scopes include multiple paragraphs) or
{\it specific} (whose answer scopes include only single paragraphs).
Table~\ref{tab:general-q} shows the number of general and specific questions given by Hybrid and human workers, respectively. For Hybrid, we only show the numbers for ``correct'' questions (i.e., those that are answered by CLIC-pages
and thus have answer scopes).
 
\begin{table}[h]
\begin{center}
\caption{Number of general and specific questions given by Hybrid and human workers (100 CLIC-page sample).}%
\begin{tabular}{|c|r|r|}
\hline
Method: & MGQs (Hybrid) & HCQs \\ \hline
Total number of questions: & 3,362 & 2,685 \\ \hline
Number of general questions: & 564 (16.8\%)  & 272 (10.1\%)  \\ \hline
Number of specific questions: & 2,798 (83.2\%)  & 2,413 (89.9\%)  \\ \hline
\end{tabular}
\label{tab:general-q}
\end{center}
\end{table}

From the table, we see that we can obtain twice as many general questions from machine than from human workers
(564 vs 272). Also, almost 90\% of all HCQs are specific ones whose answers are given by single paragraphs.
The reason why human workers tend to give predominately specific questions is that human workers generally
have short attention span; When they compose questions for a CLIC-page, they would usually
read a paragraph, understand the paragraph's content, and then think of questions that they can ask for the paragraph.
After that, they would proceed to the next paragraph and focus their attention on that. The answers to most HCQs are 
therefore paragraph specific.
From this analysis, we see that the machine is more capable of creating more general questions for the LQB.

%% file: 6-application.tex
\section{The CLIC Recommender (CRec)}
\label{sec:crec}

In the last section, we discussed how an LQB can be created with machine (by executing GPT-3 prompted by Hybrid) and also manually
with human workers. Once an LQB is created, the final component we need to bridge the legal knowledge gap (see Figure~\ref{fig:overview}) is the CLIC Recommender (CRec). 
The objective of CRec is to retrieve a short list of model questions from the LQB that are most relevant to a user's legal
situation. Instead of toiling through the large number of questions in the LQB, a user only needs to examine a small
number of model questions recommended by CRec. 
In this section, we briefly discuss CRec's design and demonstrate it with an example.

\begin{figure}[!t]
    \centering
    \includegraphics[width=0.9\textwidth]{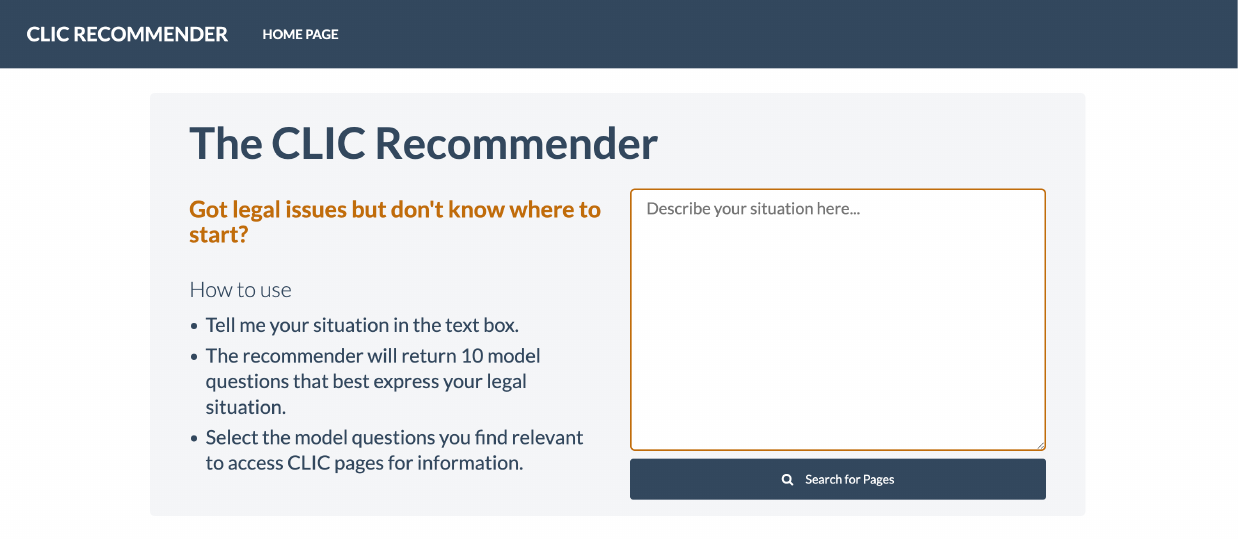}
    \caption{CRec Interface.}
    \label{fig:crec}
\end{figure}

Figure~\ref{fig:crec} shows CRec's interface. A user can describe his/her legal situation in a text box. 
Then, CRec will match the user's verbal description against the model questions in the LQB. 
This matching is done in the following way. 
First, we embed the user input text, $t_u$, into an embedding vector $v(t_u)$\footnote{We use the model all-mpnet-base-v2 to perform the text embedding.}.
Also, for each question $q \in$ LQB, we compute two embedding vectors:
\begin{itemize}
\item An {\it answer vector}, $v_a(q)$: We consider the CLIC-page paragraphs of $q$'s answer scope, i.e., $s(q).\pidlist$.
We retrieve the paragraphs' text and embed them into a single embedding vector $v_a(q)$. 
Essentially, $v_a(q)$ encodes the semantics of the answer to the question $q$.
\item A {\it question string vector}, $v_s(q)$: We embed the textual string of question $q$ into an embedding vector $v_s(q)$. This vector encodes the semantics of the question $q$.
\end{itemize}
Note that all the questions' embedding vectors ($v_a(q)$'s and $v_s(q)$'s) are pre-computed and are stored in CRec's system. 

\FloatBarrier

Next, we compute the cosine similarity between $v(t_u)$ against all the questions' answer vectors $v_a(q)$'s. 
Questions in the LQB are ranked based on their similarities against $v(t_u)$, the most similar first. 
For questions in the result that have the same answer scope (and thus the same answer vector), 
we remove all but one of them. The reason is that these questions share the same answer and thus are highly similar.
They are therefore considered redundant and only one of them should be displayed to the user in order for CRec
to provide a diversified list of model questions.  
We use the cosine similarity between questions' string vectors $v_s(q)$'s and $v(t_u)$ as tie-breakers.
Specifically, among redundant questions in the answer, the one with the highest cosine similarity between its $v_s(q)$
and $v(t_u)$ is retained, others are discarded. 
Finally, the top-10 questions in the LQB with the highest ranking (after redundancy removal) are displayed to the user.

\begin{figure}
	\centering
	\includegraphics[width=0.9\textwidth]{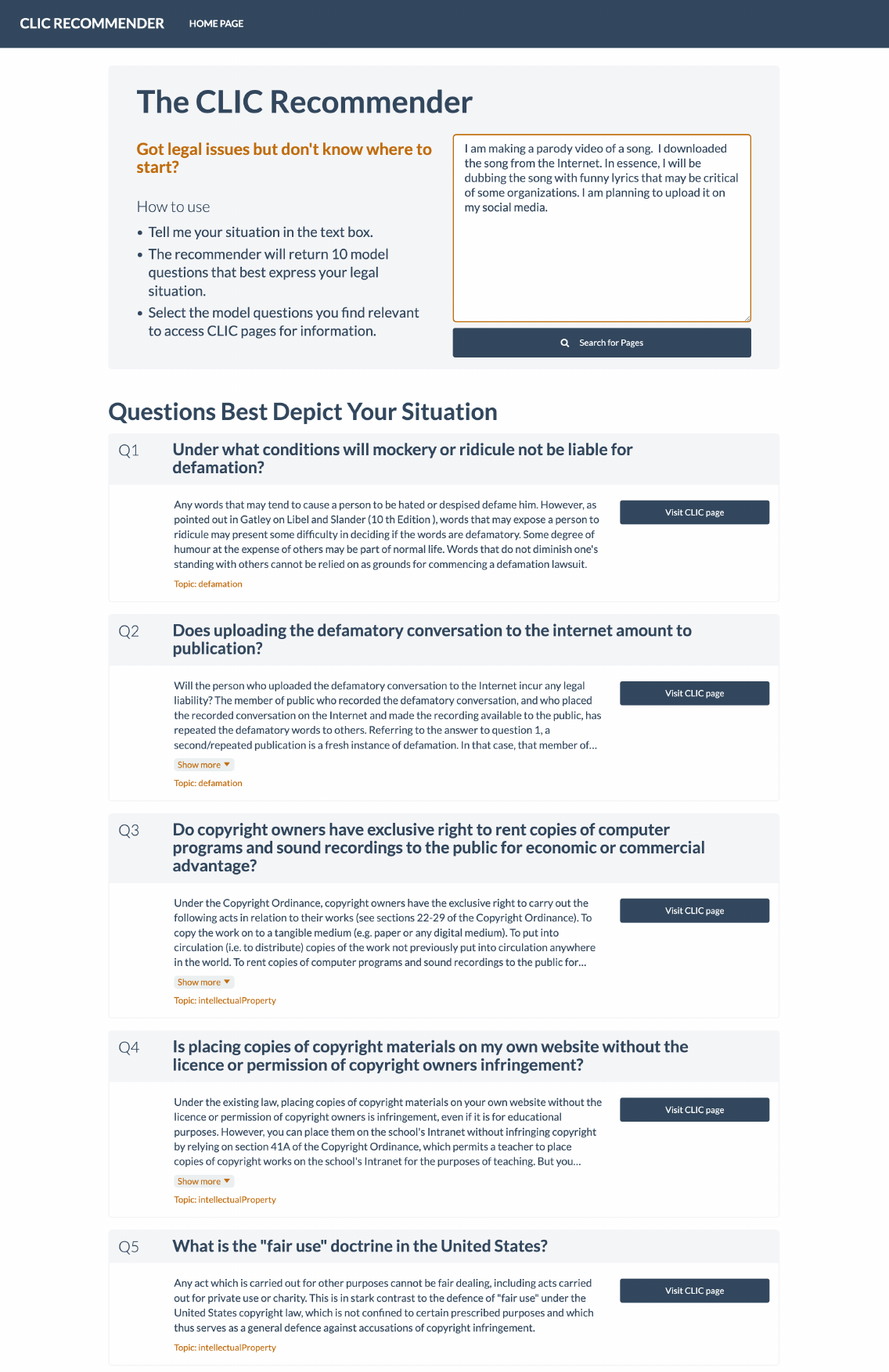}
	\caption{Recommendation results (only top 5 questions are shown due to space constraint).}
	\label{fig:crec-results}
\end{figure}

Figure~\ref{fig:crec-results} shows an example query given to CRec and the recommendation results. 
In this example, the user describes a situation where he/she intends to create a cover version of a song with mocking 
lyrics and upload the song to a social media platform. Note that in this example, the user does not explicitly ask any legal
question in the input. 
Upon receiving the user's verbal description,
CRec shortlists 10 model questions from the LQB based on the procedure we described above. 
In Figure~\ref{fig:crec-results}, we show the screenshot with the top-5 questions displayed.
In the result page, for each shortlisted question, 
CRec displays an excerpt of the answer to the question with a button on the right (labeled ``Visit CLIC page'').
The user can then go through the shortlisted questions and decide which ones are most relevant to his/her enquiry.
From Figure~\ref{fig:crec-results}, we see that many of the model questions CRec recommends (particularly questions 1, 2, 4)
are highly relevant to the user's situation.
The user can click the associated buttons next to the questions to access the full CLIC-pages for more legal information. 

We have experimented with CRec using many case scenarios found in a local public discussion forum. Our study shows that CRec is very effective in locating relevant model questions from the LQB that relate user's scenarios to legal information accessible on the CLIC platform. Our three-step approach in bridging the legal knowledge gap is therefore a promising approach in bringing legal knowledge to the public.

\FloatBarrier

%% file: 7-conclusion.tex
\section{Conclusion}
\label{sec:conclusion}
In this paper we address the legal knowledge gap, which is a barrier between formal legal sources and the general public. 
In particular, we seek to tackle the issues of navigability and comprehensibility with a focus on the former, which is to identify the legal issue in a situation faced by a layperson and locate the relevant legal concept, rule or principle in documents presenting legal information. 
We propose a three-step approach to bridging the gap. The three steps include (1) presenting and explaining legal
concepts in less technical and more readable writing, (2) creating a massive legal question bank (LQB), and (3) creating a machine legal assistant that translates a user's verbal description of a legal situation into a short list of model questions, and subsequently leads the 
user to relevant legal knowledge. We achieve the three steps by designing and implementing the CLIC platform, an LQB, and the CLIC Recommender (CRec), respectively. We perform an extensive study on the technical aspects of creating
the LQB using the large-scale pre-trained language model, GPT-3. We study how GPT-3 should be prompted to generate 
machine-generated questions (MGQs). Our finding shows that the Hybrid method performs the best in terms of a number of desirable 
properties of the LQB. These include high quantity, high precision, high coverage, and high diversity. 
The MGQs also compare favorably against human-composed questions (HCQs). 
Among the advantages, MGQs are generally more diversified than HCQs. There are also more augmenting questions created by
the machine than by human workers.
We also discuss the design of CRec and show with an example how the LQB is practically used to help a user locate relevant 
legal knowledge. 
With the three-step approach, we have demonstrated how to effectively overcome the challenge of navigability and comprehensibility, and thereby narrow the legal knowledge gap.

\section*{Statements and Declarations}
This project is supported by Innovation and Technology Fund (ITS/234/20) and the WYNG Foundation (HKU KG210018).

%% file: appendix.tex

\section{Questions created by human workers vs those composed by Machine}
\label{sec:appendix} 

\newcommand{\Hu}[3]{\item[-] [#1]: ``{#2}'' {(#3)}}

\newcommand{\Hyb}[3]{\item[-] [#1]: ``{#2}'' {(#3)}}

\newcommand{\Par}[3]{\item[-] [#1]: ``{#2}'' {(#3)}}

\newcommand{\Sec}[3]{\item[-] [#1]: ``{#2}'' {(#3)}}

\newcommand{\clicpageQ}[3]{\bigskip
	
	\noindent\hrulefill
	
	\bigskip
	
	\noindent Page $p_{#1}$: ``#2'' (\url{#3}):
	
	\bigskip\noindent}

We show the questions generated by human workers and those by various machine methods for CLIC-page 
$p_3$: ``Repairing Maintenance Obligations'' (\url{https://www.clic.org.hk/en/topics/landlord_tenant/repairing_MaintenanceObligations/q1}).
The answer scope of a question is shown in parentheses at the end of the question. For questions generated by 
Machine that are {\it incorrect}, their answer scopes are marked ``N.A.'' (for ``No Answer'').
\\


\noindent
{\bf [Questions composed by human workers]}:
\begin{itemize}
	\tiny
		\Hu{Q01}{In general, how do parties specify maintenance responsibility in tenancy agreements?}{$p_3$:[heading]}
		\Hu{Q02}{Who should be responsible for external and structural maintenance?}{$p_3$:[2]}
		\Hu{Q03}{Who should be responsible for internal and non-structural maintenance?}{$p_3$:[2]}
		\Hu{Q04}{What kind of maintenance should be responsible by the landlord?}{$p_3$:[2]}
		\Hu{Q05}{What kind of maintenance should be responsible by the tenant?}{$p_3$:[2]}
		\Hu{Q06}{Are the terms like external, internal, structural, or non-structural clear enough to specify maintenance responsibility?}{$p_3$:[2, 3]}
		\Hu{Q07}{Is it fair for a tenant to become responsible for property maintenance?}{$p_3$:[4]}
		\Hu{Q08}{Do tenants have the responsibility to maintain the property from "fair wear and tear"?}{$p_3$:[5]}
		\Hu{Q09}{Without the notice of structural defects, is the landlord responsible for structural repair?}{$p_3$:[5]}
		\Hu{Q10}{Is it reasonable for the landlord to be notified before being responsible for structural repair?}{$p_3$:[5]}
		\Hu{Q11}{Why some landlord may volunteer to carry out repairs and maintenance works?}{$p_3$:[7]}
		\Hu{Q12}{What are the advantages for landlords to repair and maintain the property voluntarily?}{$p_3$:[7]}
		\Hu{Q13}{If tenants become incooperative to landlord carrying out necessary inspection and repair works, what can the landlord do?}{$p_3$:[8]}
\end{itemize}

\bigskip\noindent
{\bf [Questions generated by Hybrid]}:
\begin{itemize}
\tiny
\Hyb {Q01} {What should I do if I have a problem with my property that needs to be repaired or maintained?}{N.A.}
\Hyb {Q02} {Who is responsible for external and structural repairs and maintenance?}{$p_3$:[2]}
\Hyb {Q03} {Who is responsible for internal and non-structural repairs and maintenance?}{$p_3$:[2]}
\Hyb {Q04} {What is the meaning of "fair wear and tear excepted"?}{$p_3$:[5]}
\Hyb {Q05} {What is the landlord's obligation for structural repairs and maintenance?}{$p_3$:[2]}
\Hyb {Q06} {How can I find out who is responsible for repairs and maintenance?}{$p_3$:[1]}
\Hyb {Q07} {Can the landlord volunteer to carry out repairs and maintenance works?}{$p_3$:[7]}
\Hyb {Q08} {What if the tenant becomes uncooperative?}{$p_3$:[8]}
\Hyb {Q09} {Can the landlord apply for an injunction?}{$p_3$:[8]}
\Hyb {Q10} {Can the landlord terminate the tenancy agreement?}{$p_3$:[8]}
\Hyb {Q11} {What is the difference between external and internal repairs?}{$p_3$:[2]}
\Hyb {Q12} {What is the difference between structural and non-structural repairs?}{$p_3$:[2]}
\Hyb {Q13} {Who is responsible for repair and maintenance of the property?}{$p_3$:[heading]}
\Hyb {Q14} {What is the difference between "fair wear and tear" and "damage"?}{$p_3$:[5]}
\Hyb {Q15} {What is the landlord's obligation for structural repairs and maintenance?}{$p_3$:[2]}
\Hyb {Q16} {What is the tenant's obligation for repair and maintenance of the property?}{$p_3$:[heading]}
\Hyb {Q17} {Can the landlord carry out repair and maintenance works without the tenant's consent?}{$p_3$:[8]}
\Hyb {Q18} {What if the tenant refuses to allow the landlord to carry out repair and maintenance works?}{$p_3$:[8]}
\Hyb {Q19} {Can the landlord terminate the tenancy agreement if the tenant refuses to allow him/her to carry out repair and maintenance works?}{$p_3$:[8]}
\Hyb {Q20} {What can the tenant do if he/she is not satisfied with the repair and maintenance works carried out by the landlord?}{N.A.}
\Hyb {Q21} {What is a typical approach under tenancy agreements when it comes to repair and maintenance responsibilities?}{$p_3$:[2]}
\Hyb {Q22} {How can a simple dichotomy between landlord and tenant responsibilities for repair and maintenance be problematic?}{$p_3$:[2]}
\Hyb {Q23} {What should a well-drafted tenancy document do in order to clarify the parties' duties with regards to repair and maintenance?}{$p_3$:[3]}
\Hyb {Q24} {Is it common for tenants to be responsible for onerous duties such as repair and maintenance under their tenancy agreement?}{$p_3$:[4]}
\Hyb {Q25} {What is the significance of the phrase "fair wear and tear excepted" with regards to a tenant's obligations for repair and maintenance?}{$p_3$:[5]}
\Hyb {Q26} {What is the role of notice clauses in specifying a landlord's obligations for structural repairs and maintenance?}{$p_3$:[5]}
\Hyb {Q27} {Why might landlords agree to incur expenses to remedy defects in the property even though they are not contractually obliged to do so?}{$p_3$:[7]}
\Hyb {Q28} {In what circumstances might a landlord exercise a right to enter the property and carry out necessary inspection and repair works?}{$p_3$:[8]}
\Hyb {Q29} {What could happen if a tenant becomes uncooperative when a landlord tries to exercise such a right?}{$p_3$:[8]}
\Hyb {Q30} {Is it possible for a landlord to terminate a tenancy agreement if a tenant does not cooperate with repair and maintenance works?}{$p_3$:[8]}
\Hyb {Q31} {What is the fair wear and tear exception?}{$p_3$:[5]}
\Hyb {Q32} {What does it mean when a tenancy agreement states that the tenant's obligations for repair and maintenance are limited by the phrase "fair wear and tear excepted"?}{$p_3$:[5]}
\Hyb {Q33} {What is the landlord's obligation for structural repairs and maintenance?}{$p_3$:[2]}
\Hyb {Q34} {What is the tenant's obligation for repair and maintenance?}{$p_3$:[heading]}
\Hyb {Q35} {How can I find out who is responsible for repairs and maintenance?}{$p_3$:[1]}
\Hyb {Q36} {How can I find out if the landlord is responsible for structural repairs and maintenance?}{$p_3$:[1, 2, 6]}
\Hyb {Q37} {How can I find out if the tenant is responsible for repair and maintenance?}{$p_3$:[1, 2, 6]}
\Hyb {Q38} {Is it common for a tenant to be responsible for repair and maintenance?}{$p_3$:[4]}
\Hyb {Q39} {What is the landlord's right to enter the property and carry out necessary inspection and repair works?}{$p_3$:[8]}
\Hyb {Q40} {Can the landlord terminate the tenancy agreement if the tenant is uncooperative?}{$p_3$:[8]}
\Hyb {Q41} {What is the fair wear and tear exception?}{$p_3$:[5]}
\Hyb {Q42} {What does it mean when a tenant's obligations for repair and maintenance are limited by the phrase "fair wear and tear excepted"?}{$p_3$:[5]}
\Hyb {Q43} {What is included in the fair wear and tear exception?}{$p_3$:[5]}
\Hyb {Q44} {How is the fair wear and tear exception determined?}{N.A.}
\Hyb {Q45} {What are some examples of fair wear and tear?}{N.A.}
\Hyb {Q46} {Is there a limit to the amount of damage that can be caused by fair wear and tear?}{N.A.}
\Hyb {Q47} {What if the tenant causes damage that is not considered to be fair wear and tear?}{N.A.}
\Hyb {Q48} {Can the landlord repair or replace items that are damaged by fair wear and tear?}{$p_3$:[5, 7]}
\Hyb {Q49} {Who is responsible for repairing or replacing items that are damaged by fair wear and tear?}{$p_3$:[5, 7]}
\Hyb {Q50} {What if the landlord and tenant cannot agree on what is considered to be fair wear and tear?}{N.A.}
\Hyb {Q51} {What is the "fair wear and tear" exception in a tenancy agreement?}{$p_3$:[5]}
\Hyb {Q52} {What is the landlord's obligation for structural repairs and maintenance?}{$p_3$:[2]}
\Hyb {Q53} {How can I tell if the damage is due to "fair wear and tear" or not?}{$p_3$:[5]}
\Hyb {Q54} {I am a tenant. Can I carry out repair and maintenance works on my own?}{$p_3$:[4]}
\Hyb {Q55} {I am a landlord. Can I enter the property to carry out repair and maintenance works without the tenant's permission?}{$p_3$:[8]}
\Hyb {Q56} {My tenancy agreement does not contain any clauses regarding repair and maintenance. Who is responsible for repairs and maintenance?}{$p_3$:[heading]}
\Hyb {Q57} {My tenancy agreement contains a clause stating that the tenant is responsible for all repair and maintenance works. Is this fair?}{$p_3$:[1, 4]}
\Hyb {Q58} {My tenancy agreement contains a clause stating that the landlord is responsible for all repair and maintenance works. Is this fair?}{$p_3$:[1, 7]}
\Hyb {Q59} {I am a tenant. The property I am renting is in a state of disrepair. What can I do?}{$p_3$:[heading]}
\Hyb {Q60} {I am a landlord. The property I am renting out is in a state of disrepair. What can I do?}{$p_3$:[heading]}
\Hyb {Q61} {Can the landlord enter the property without the tenant's permission?}{$p_3$:[8]}
\Hyb {Q62} {What if the tenant refuses to allow the landlord to enter the property?}{$p_3$:[8]}
\Hyb {Q63} {What if the tenant is not cooperative?}{$p_3$:[8]}
\Hyb {Q64} {What if the landlord needs to carry out urgent repairs?}{$p_3$:[8]}
\Hyb {Q65} {What if the tenancy agreement does not impose any duty on the landlord?}{$p_3$:[6, 7, 8]}
\Hyb {Q66} {Can the landlord terminate the tenancy agreement?}{$p_3$:[8]}
\Hyb {Q67} {What if the tenant objects to the repairs?}{$p_3$:[8]}
\Hyb {Q68} {Can the landlord charge the tenant for the cost of repairs?}{N.A.}
\Hyb {Q69} {What if the repairs are not carried out properly?}{N.A.}
\Hyb {Q70} {Can the tenant sue the landlord for breach of contract?}{N.A.}
\Hyb {Q71} {What is the definition of "serious water leakage"?}{N.A.}
\Hyb {Q72} {What is the landlord's right to enter the property?}{$p_3$:[8]}
\Hyb {Q73} {What is the landlord's duty to carry out necessary inspection and repair works?}{$p_3$:[8]}
\Hyb {Q74} {What is the tenant's obligation to cooperate with the landlord?}{$p_3$:[8]}
\Hyb {Q75} {What are the consequences if the tenant does not cooperate with the landlord?}{$p_3$:[8]}
\Hyb {Q76} {Can the landlord terminate the tenancy agreement if the tenant does not cooperate?}{$p_3$:[8]}
\Hyb {Q77} {What is an interlocutory injunction?}{N.A.}
\Hyb {Q78} {How does an interlocutory injunction work?}{N.A.}
\Hyb {Q79} {What are the consequences of an interlocutory injunction?}{$p_3$:[8]}
\Hyb {Q80} {Is there any other way for the landlord to obtain possession of the property besides an interlocutory injunction?}{$p_3$:[8]}
\end{itemize}

\bigskip\noindent
{\bf [Questions generated by Paragraph-based]}:
\begin{itemize}
\tiny
\Par {Q01} {What is the Residential Tenancies Act?}{N.A.}
\Par {Q02} {What is a tenancy agreement?}{N.A.}
\Par {Q03} {What are the landlord’s responsibilities?}{$p_3$:[heading]}
\Par {Q04} {What are the tenant’s responsibilities?}{$p_3$:[heading]}
\Par {Q05} {Can the landlord enter the rented premises?}{$p_3$:[8]}
\Par {Q06} {Can the tenant be evicted?}{N.A.}
\Par {Q07} {What is rent in advance?}{N.A.}
\Par {Q08} {Can the landlord increase rent?}{N.A.}
\Par {Q09} {Can the tenant sublet or assign the tenancy?}{N.A.}
\Par {Q10} {What happens if there is a dispute between the landlord and tenant?}{N.A.}
\Par {Q11} {What is the difference between an internal and external repair?}{$p_3$:[2]}
\Par {Q12} {What is the difference between a structural and non-structural repair?}{$p_3$:[2]}
\Par {Q13} {Who is responsible for the maintenance of common areas?}{N.A.}
\Par {Q14} {Who is responsible for the upkeep of gardens and landscaping?}{N.A.}
\Par {Q15} {Who pays for water and electricity consumption?}{N.A.}
\Par {Q16} {What about Council rates and taxes?}{N.A.}
\Par {Q17} {What insurance cover is required?}{N.A.}
\Par {Q18} {What happens if the property is damaged by a natural disaster?}{N.A.}
\Par {Q19} {What are the responsibilities of the landlord in relation to health and safety?}{N.A.}
\Par {Q20} {Can the landlord make changes to the property without the tenant's consent?}{$p_3$:[8]}
\Par {Q21} {What is a tenancy agreement?}{N.A.}
\Par {Q22} {What should be included in a tenancy agreement?}{N.A.}
\Par {Q23} {How long does a tenancy agreement last?}{N.A.}
\Par {Q24} {What is a security deposit?}{N.A.}
\Par {Q25} {When is the security deposit returned?}{N.A.}
\Par {Q26} {Who is responsible for repairing damage to the property?}{$p_3$:[heading]}
\Par {Q27} {Who is responsible for paying utilities?}{N.A.}
\Par {Q28} {Can a landlord enter the property without notice?}{$p_3$:[8]}
\Par {Q29} {Can a tenant sublet the property?}{N.A.}
\Par {Q30} {What is a lease?}{N.A.}
\Par {Q31} {How long does a lease last?}{N.A.}
\Par {Q32} {What is included in a lease?}{N.A.}
\Par {Q33} {Can a landlord evict a tenant?}{N.A.}
\Par {Q34} {Can a tenant break a lease?}{N.A.}
\Par {Q35} {What is rent?}{N.A.}
\Par {Q36} {How much rent can a landlord charge?}{N.A.}
\Par {Q37} {Can a landlord raise rent?}{N.A.}
\Par {Q38} {What is security deposit?}{N.A.}
\Par {Q39} {Can a landlord keep my security deposit?}{N.A.}
\Par {Q40} {What is a tenancy agreement?}{N.A.}
\Par {Q41} {What is the difference between a tenancy agreement and a lease?}{N.A.}
\Par {Q42} {How long does a tenancy agreement last?}{N.A.}
\Par {Q43} {What is the difference between a periodic tenancy and a fixed-term tenancy?}{N.A.}
\Par {Q44} {What is the difference between a verbal tenancy agreement and a written tenancy agreement?}{N.A.}
\Par {Q45} {Can a landlord change the terms of a tenancy agreement?}{N.A.}
\Par {Q46} {Can a tenant change the terms of a tenancy agreement?}{N.A.}
\Par {Q47} {What are the landlord's obligations under a tenancy agreement?}{$p_3$:[2, 6]}
\Par {Q48} {What are the tenant's obligations under a tenancy agreement?}{$p_3$:[2, 6]}
\Par {Q49} {Who is responsible for maintaining and repairing the property?}{$p_3$:[heading]}
\Par {Q50} {How long does a tenancy agreement last?}{N.A.}
\Par {Q51} {How much rent can a landlord charge?}{N.A.}
\Par {Q52} {How often can a landlord increase the rent?}{N.A.}
\Par {Q53} {What amenities and services must a landlord provide?}{N.A.}
\Par {Q54} {Can a tenant make changes to the property?}{N.A.}
\Par {Q55} {Can a landlord enter the property?}{$p_3$:[8]}
\Par {Q56} {What happens if there is damage to the property?}{$p_3$:[heading]}
\Par {Q57} {What happens if there are problems with neighbours?}{N.A.}
\Par {Q58} {What happens if I want to move out?}{N.A.}
\Par {Q59} {What happens if I want my landlord to move out?}{N.A.}
\Par {Q60} {What is the landlord's responsibility for maintaining and repairing the property?}{$p_3$:[heading]}
\Par {Q61} {Who is responsible for maintaining and repairing the property if the tenant and landlord disagree?}{$p_3$:[heading]}
\Par {Q62} {What if the landlord volunteers to carry out repairs and maintenance works?}{$p_3$:[7]}
\Par {Q63} {How can I ensure that my tenancy agreement covers all potential repair and maintenance costs?}{N.A.}
\Par {Q64} {What if there is damage to the property caused by a tenant?}{$p_3$:[5]}
\Par {Q65} {Can a tenant be held responsible for preexisting damage to the property?}{$p_3$:[5]}
\Par {Q66} {How can I get my deposit back if there is damage to the property?}{N.A.}
\Par {Q67} {What if I want to make changes to the property?}{N.A.}
\Par {Q68} {Can I terminate my tenancy agreement early if the property is not maintained or repaired?}{N.A.}
\Par {Q69} {Who do I contact if I have a problem with my landlord or property?}{N.A.}
\Par {Q70} {What is the Landlord and Tenant (Consolidation) Ordinance (Cap. 7)?}{N.A.}
\Par {Q71} {What are the rights and duties of landlords and tenants?}{$p_3$:[heading]}
\Par {Q72} {How can I renew my tenancy agreement?}{N.A.}
\Par {Q73} {What are the terms of a standard tenancy agreement?}{$p_3$:[2]}
\Par {Q74} {Can I sublet my property?}{N.A.}
\Par {Q75} {Can the landlord enter my property without my permission?}{$p_3$:[8]}
\Par {Q76} {Can I change the locks on my property?}{N.A.}
\Par {Q77} {Can I be evicted from my property?}{N.A.}
\Par {Q78} {Can I withhold rent if the property is not in good condition?}{N.A.}
\Par {Q79} {In general, who shall be responsible for maintaining and repairing the property?}{$p_3$:[heading]}

\end{itemize}

\bigskip\noindent
{\bf [Questions generated by Section-based]}:
\begin{itemize}
\tiny
	\Sec{Q01}{What is the landlord’s duty to repair?}{$p_3$:[heading]}
	\Sec{Q02}{What is the tenant’s duty to repair?}{$p_3$:[heading]}
	\Sec{Q03}{How can I make sure that I will not be held liable for any damage caused by me, my family or visitors to the property?}{$p_3$:[5]}
	\Sec{Q04}{What should I do if I find that there are defects in the property which need to be repaired?}{$p_3$:[heading]}
	\Sec{Q05}{My rental flat is infested with cockroaches/bedbugs/mice. What can I do?}{$p_3$:[heading]}
	\Sec{Q06}{There is water leakage in my rental flat. What can I do?}{$p_3$:[heading]}
	\Sec{Q07}{The lift in my building has broken down and I am unable to use it. What can I do?}{N.A.}
	\Sec{Q08}{There is no hot water supply in my rental flat. What can I do?}{$p_3$:[heading]}
	\Sec{Q09}{The air-conditioner in my rental flat has broken down and I am unable to use it. What can I do?}{$p_3$:[heading]}
	\Sec{Q10}{Can I withhold rent if my landlord fails to carry out repair works?}{N.A.}

\end{itemize}